\pdfoutput=1

\documentclass[11pt]{article}

\usepackage[preprint]{acl}

\usepackage{times}
\usepackage{latexsym}

\usepackage[T1]{fontenc}

\usepackage[utf8]{inputenc}

\usepackage{microtype}

\usepackage{inconsolata}

\usepackage{graphicx}
\usepackage{amsmath}
\usepackage{amsthm}
\usepackage{amssymb}
\usepackage{array}
\usepackage{booktabs}
\usepackage{makecell}
\usepackage{multirow}
\usepackage{algorithm}
\usepackage{algorithmic}
\usepackage{enumitem}
\usepackage[breakable]{tcolorbox}
\usepackage{xcolor}
\usepackage{supertabular,booktabs}
\definecolor{light-gray}{gray}{0.9}
\usepackage{ltxtable}
\usepackage{longtable}
\usepackage{pifont}
\usepackage{amssymb}
\usepackage{wasysym}
\usepackage{subfigure}
\usepackage{tabularx}
\usepackage{authblk}
\usepackage{ragged2e}
\usepackage{authblk}

\title{Don't Half-listen: Capturing Key-part Information in Continual Instruction Tuning}

\author{
    \textbf{Yongquan He}$^{2\dag}$\textbf{, }
    \textbf{Wenyuan Zhang}$^{4\dag}$\textbf{, }  
    \textbf{Xuancheng Huang}$^{3\dag}$\textbf{, } 
    \textbf{Peng Zhang}$^{1*}$\textbf{, } 
    \\
    \textbf{Lingxun Meng}$^{2}$\textbf{, } 
    \textbf{Xiang Zhou}$^{2}$\textbf{, }  
    \textbf{Ke Zeng}$^{2}$\textbf{, } 
    \textbf{Xunliang Cai}$^{2}$\\ 
    $^1$School of Cyber Science and Engineering, Nanjing University of Science and Technology, China \\
    $^2$Meituan, China \\
    $^3$Zhipu.AI, China \\
    $^4$Institute of Information Engineering, Chinese Academy of Sciences, China \\
    \texttt{heyongquan\{@meituan, 1997@gmail\}.com}\\
}

\begin{document}
\maketitle

\begin{abstract}

{\let\thefootnote\relax
\footnotetext{
$\dag$ \ Equal contribution.
}}

{\let\thefootnote\relax
\footnotetext{
* \ Corresponding authors: Peng Zhang (\url{zhang_peng@njust.edu.cn}).
}}

Instruction tuning for large language models (LLMs) can drive them to produce results consistent with human goals in specific downstream tasks.
However, the process of continual instruction tuning (CIT) for LLMs may bring about the catastrophic forgetting (CF) problem, where previously learned abilities are degraded.
Recent methods try to alleviate the CF problem by modifying models or replaying data, which may only remember the surface-level pattern of instructions and get confused on held-out tasks.
In this paper, we propose a novel continual instruction tuning method based on Key-part Information Gain (\textbf{KPIG}).
Our method computes the information gain on masked parts to dynamically replay data and refine the training objective, which enables LLMs to capture task-aware information relevant to the correct response and alleviate overfitting to general descriptions in instructions.
In addition, we propose two metrics, \texttt{P-score} and \texttt{V-score}, to measure the generalization and instruction-following abilities of LLMs.
Experiments demonstrate our method achieves superior performance on both seen and held-out tasks.
\end{abstract}

\section{Introduction}
Large language models (LLMs) make remarkable breakthroughs in recent years \cite{llmsurvey}.
LLMs such as PaLM \cite{palm} and LLaMA \cite{llama} show powerful capabilities in multiple tasks such as information extraction, question answering, commonsense reasoning, and mathematical operations.
One of the major issues is how to leverage the knowledge of LLMs pretrained with unsupervised or general objectives to produce results consistent with human intent during task-specific interactions \cite{itsurvey}.
To endow LLMs with such ``instruction-following'' ability, instruction tuning is proposed as an effective technique that can bridge the gap between the generation process of LLMs and the objective of users \cite{instructgpt, itsurvey, citsurvey}.

\begin{figure}
    \centering
    \includegraphics[width=7.5cm]{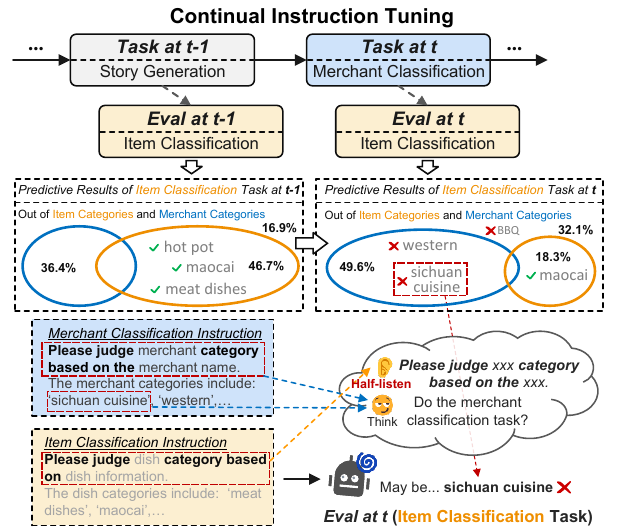}
    \caption{Task confusion on item classification (IC) after training merchant classification (MC). Note that IC is a held-out task for evaluation, and LLM at $t$ generates more illegal categories defined in MC ($36.4\% \rightarrow 49.6\%$) as their instructions are similar.
}
    \label{fig:motivation}
\end{figure}

Although tuning a pretrained LLM with instruction data before deployment gains wide application, 
it still faces challenges when dealing with incremental data and tasks \cite{citb}.
Continual learning (CL) \cite{cl} is introduced to avoid costly retraining on all collected instances \cite{cl}, and continual instruction tuning (CIT) \cite{citb} is a sub-task of it about instruction data.
However, catastrophic forgetting (CF) is still an unavoidable problem during CIT, which refers to the forgetting of previously learned tasks and the deterioration of original generalization ability \cite{pcll, dcl, citb}.

Recently, \textit{replay}, \textit{architecture}, and \textit{regularization} are three main strategies to mitigate the CF problem.
\textit{Replay} is the most prevalent strategy that leverages task-specific features to replay a small set of previous data \cite{contintin, dynainst} or generated pseudo samples \cite{pcll, dcl}.
\textit{Architecture} obtains the target model by performing a model merging of other available LLMs \cite{lmcocktail, dare} or introducing task-specific components for newly emerging tasks \cite{adapter, lora}.
Moreover, \textit{regularization} is usually utilized as a penalty strategy to alleviate overfitting on seen tasks \cite{ewc}.

Despite their impressive performance on seen tasks, these methods may only learn surface-level patterns \cite{itsurvey} of instructions when applied to the CIT scenario.
This observation is supported by prior research \cite{prove}, which suggests that LLMs may generate unchanged responses on seen tasks and become confused on held-out tasks, even if we modify some components in original instructions.
We also observe a similar phenomenon, as shown in Figure~\ref{fig:motivation}, compared to responses of the item classification task at $t-1$, the LLM generates more illegal categories that are not defined in the item classification instruction at $t$ (after training on the merchant classification task).
This half-listening phenomenon indicates that the overfitting to seen instructions is serious in CIT, potentially leading to confusion during inferring on held-out tasks. 
Therefore, we focus on a new challenge concerning the degradation of instruction-following and generalization abilities within the CIT framework, both of which are essential abilities of instruction-based LLMs.

In this paper, we propose a novel CIT paradigm based on key-part information gain (\textbf{KPIG}) to handle the above challenge.
\textbf{Key parts} are consecutive spans in the instruction which provide task-aware guidance on the content, length, and format to generate desired responses.
And we expect that LLMs can be sensitive to key parts for task-aware performance, which exhibits strong instruction-following and generalization abilities on various tasks.
Firstly, we rewrite the instructions and corresponding key parts to diversify the combination of key parts and general descriptions.
Then we selectively replay a small set of historical data whose information gain (IG) is the lowest.
And IG is our proposed indicator used to measure the task-aware ability of LLMs, which is calculated by masking the key parts.
Finally, we apply a Jensen–Shannon \cite{jsd} divergence (JSD) on masked instructions, and IG is utilized as a dynamic temperature, to increase the IG margin relative to the surface-level patterns.
Moreover, as instruction-following and generalization abilities are our concerns, we propose two novel evaluation metrics, \texttt{P-score} and \texttt{V-score}, instead of simply using Rouge-L \cite{rouge} as previous methods \cite{dynainst, citb}.
Experiments conducted on Super-NaturalInstructions (SupNatInst) \cite{supernatural} and our Chinese domain (Domain) datasets show superior performance on both seen and held-out tasks, and violations of instructions such as out-of-scope, wordy statements, and illegal formats are reduced.

Our contributions can be summarized as follows:
1) We propose a novel CIT paradigm by masking key parts to alleviate the half-listening problem of 
 instructions.
2) We propose information gain as an indicator for measuring task-aware ability, which serves to dynamically replay data and refine the training objective.
3) We propose a novel evaluation metric \texttt{V-score} centered on instruction-following ability. 
4) Compared to other CL baselines, our method achieves state-of-the-art performance on public and domain datasets.

\begin{figure*}[t!]
    \centering
    \includegraphics[width=0.9\linewidth]{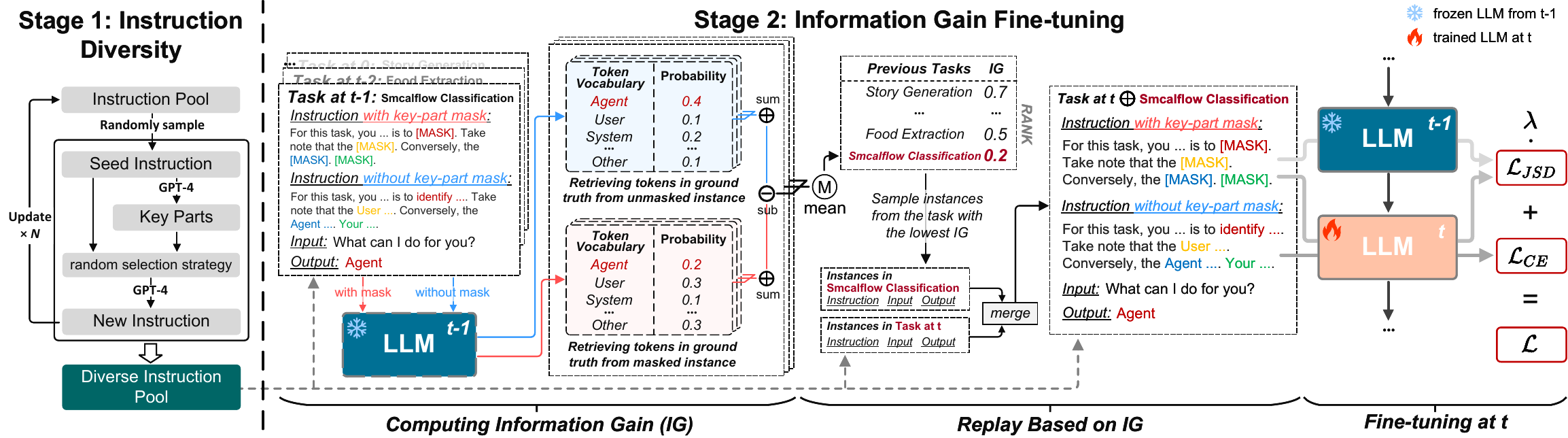}
    \caption{The continual instruction tuning framework of our KPIG. In the instruction diversity stage, we require GPT-4 to pay more attention to key parts during the rewriting process. In the information gain fine-tuning stage, we dynamically replay previous tasks with our learning objective based on IG to alleviate the half-listening problem.}
    \label{fig:framework}
\end{figure*}

\section{Related Work}
\subsection{Instruction Tuning}
LLMs show powerful emergent abilities in many downstream tasks \cite{palm, llama2, llmsurvey}.
Since most LLMs are typically pretrained with the next word prediction error on large corpora, instruction tuning is proposed as an effective technique to further enhance the instruction-following ability of the generation process \cite{instructgpt, itsurvey}.
And increasing the quantity, diversity, and creativity of instructions is empirically validated as an effective strategy to improve the instruction-following and generalization capabilities of LLMs \cite{itsurvey, wizardlm, agentuning}.
Collecting existing datasets and synthesizing data with LLMs are main strategies to obtain high-quality instruction data \cite{itsurvey, citsurvey,zhang2024revealing}.
The former collects existing data and converts it into instruction-style datasets through templates or machine translation \cite{belle, alpaca}, while the latter like Evol-Instruct \cite{wizardlm}, instructWild \cite{instructwild} and Self-Instruct \cite{selfinstruct} ask LLMs to rewrite seed instructions based on specific strategies.
In addition to increasing the diversity of task data, optimizing LLM with a comparison dataset collected by human feedback or LLMs also helps to generate desired responses \cite{instructgpt, itsurvey}.
However, LLMs may only remember surface-level patterns of seen instructions, causing the output results not satisfy all constraints on held-out instructions \cite{itsurvey, prove}.
In this paper, we propose a key-part information mask mechanism to make LLMs focus more on tokens in instructions that are pertinent to the content, length, and format of the ground truths.

\subsection{Continual Learning}
Compared with multi-task learning, CL \cite{cl} refers to learning from sequential data across multiple time steps, which may lead to CF problem.
Since CIT \cite{citb} is a sub-task of CL applied to instruction data, we do not discuss them separately.
Recent methods mainly focus on tackling the forgetting of previously learned tasks, and CITB \cite{citb} categorizes them into three groups, \textit{replay}, \textit{architecture}, and \textit{regularization}.
\textit{Replay}-based methods replay experience with historical data \cite{contintin, cit2, dynainst} or generated pseudo samples \cite{pcll, dcl}, while \textit{architecture}-based methods introduce task-specific parameters \cite{adapter, lora} or gradually merging models trained on different tasks \cite{lmcocktail, dare}.
Moreover, \textit{regularization}-based methods are strategies for objective optimization and overfitting penalty \cite{kd, ewc}, which are used alone or in combination with other methods \cite{dynainst, pcll, dcl}.
Despite effectively alleviating the forgetting of previously learned tasks, they lack attention to instructions and may half-listen to surface-level descriptions in held-out instructions when applied to CIT.
In this paper, we selectively replay a few historical data and employ temperature based on the IG of masked key parts, which encourages LLMs to be more sensitive to task-aware information in instructions.

\section{Methodology}

This section introduces our proposed method, named Key-part Information Gain (\textbf{KPIG}), for continual instruction tuning on LLMs.
We first define the task and notations in \S\ref{3.1}. 
Then we detail our instruction diversity module (\S\ref{3.2}) and information gain fine-tuning (\S\ref{3.3}) module in Figure~\ref{fig:framework}.
Moreover, considering the specificity of sequential training in CIT, we introduce how to reconstruct datasets and evaluate performance (\S\ref{3.4}).

\subsection{Task Definition and Notations}\label{3.1}
We finetune a LLM with a stream of task sets $\mathcal{T}^T = \{\tau_{1}, \tau_{2}, \ldots,\tau_{n}\}_{t=1}^T$ sequentially, where $T$ is the number of time steps and $n$ is the number of tasks at corresponding time $t$.
Each instance $d_\tau$ in the task $\tau$ can be formed as a triple $(i, c, y)$: instruction $i$, which is a natural language text to demonstrate the definition of current task in human style; an optional context $c$ which provides supplementary information for context; an expected output $y$ corresponding to the instruction and the context.
And each task $\tau$ can be split into $\tau_{train}$ and $\tau_{test}$.
At each time step $t$, we finetune the LLM on a mixture of $\tau_{train}$, where $\tau \in \mathcal{T}^t$.
After completing the $T$-step training, we evaluate its performance on the $\tau_{test}$ of seen tasks $\mathcal{T}_{seen}$ and held-out tasks $\mathcal{T}_{unseen}$, where $\mathcal{T}_{seen} \cap \mathcal{T}_{unseen} = \varnothing$.

\begin{figure}
    \centering
    \includegraphics[width=7cm]{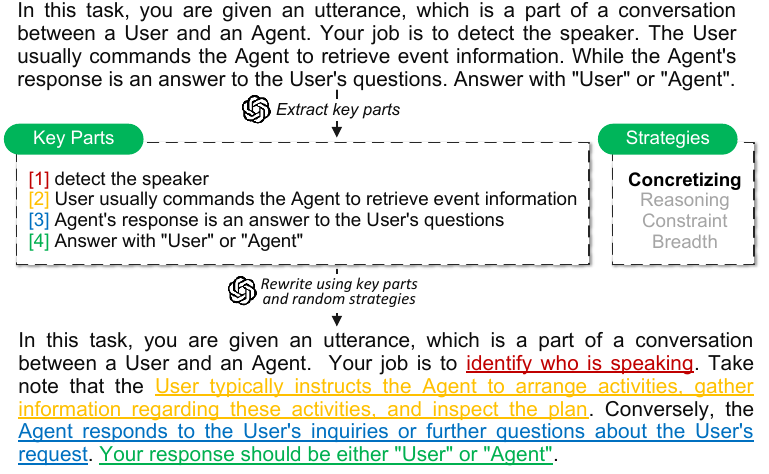}
    \caption{An example of the instruction diversity.}
    \label{fig:instruction}
\end{figure}

\subsection{Instruction Diversity}\label{3.2}

Due to the inefficiency of manual annotation, the number of instructions for a task may be scarce \cite{itsurvey,zhang2025s1benchsimplebenchmarkevaluating}.
Taking the SupNatInst as an example, each task has hundreds to thousands of instances but only one human-handwritten instruction.
Moreover, humans may struggle to produce different instructions with the same meaning \cite{wizardlm,zhang2025sotopiaomegadynamicstrategyinjection}.
Motivated by WizardLM \cite{wizardlm}, we diversify the combination of key parts and other general descriptions in instructions via GPT-4 \cite{gpt4} and different templates, which aims to prompt the LLM to identify task-aware information in instructions with varying levels of complexity.

As shown in Figure~\ref{fig:instruction}, we first ask GPT-4 to generate key parts of the seed instruction and return them as a list.
Then we input seed instructions and their corresponding key parts to gain new instructions and key parts recursively.
As we expect that key parts can play an important role in controlling text generation, we encourage GPT-4 to focus on key parts when evolving with the following strategies: 1) Concretizing, which replaces general concepts in key parts with more specific concepts. 2) Reasoning, which explicitly requests multiple-step reasoning if key parts can be organized into a few simple thinking processes. 3) Constraint, which adds one more constraint on seed instructions. 4) Breadth, which rewrites the seed instruction while keeping length close and key parts unchanged.
It should be noted that we also need to give GPT-4 some demonstrations when rewriting the instructions and obtaining the key parts.
More details about templates can be found in Appendix~\ref{template}.

\subsection{Information Gain Fine-tuning} \label{3.3}
Although diversified instructions are proven helpful to train instruction-following LLMs \cite{wizardlm}, we argue that LLMs sometimes may only half-listen to surface-level patterns \cite{itsurvey, prove}, which refers to overfitting on seen instructions and confusion with held-out tasks.
To measure the task-aware ability of current LLM for specific task $\tau$, we propose the concept of information gain (IG) by masking key parts in each instance $d_\tau = (i, c, y)$:
\begin{equation}
\mathcal{G}(d_\tau, d_\tau^m) = {\rm Info}(y|x) - {\rm Info}\big(y|{\rm mask}(x)\big),
\label{ig}
\end{equation}
where $d_\tau \in \tau$ and ${\rm Info}(\cdot)$ is a weighted sum of the generation probabilities of sequence $y$.
Meanwhile, ${\rm mask(\cdot)}$ denotes the operation of applying a key-part mask to the instruction $i$, which replaces the key parts in instruction $i$ with \texttt{[MASK]} symbol to obtain mask instruction $i^m$ and surface-level instance $d_\tau^m = (i^m, c, y)$.
Giving an input $x = (i, c)$ and an expected output $y = \{t_{1}, t_{2}, \ldots,t_{K}\}$, where $K$ is the length of $y$ after the tokenization process.
The ${\rm Info}(y|x)$ can be calculated as follows: 
\begin{equation}
{\rm Info}(y|x) = {\rm Info}\big(y|(i, c)\big) = \sum_{k=1}^{K}\alpha^k p(t_k),
\end{equation}
where $p(t_k)$ is the probability of expected token $t_k$ in $y$ given the concatenation of input $x$ and the previously generated output tokens $t_{history}$.
$\alpha$ is an exponential decay hyperparameter because the probability of subsequent tokens always becomes greater as the inference process progresses.
As for the $p(t_k)$, we gain it via retrieving from the softmax function results on the head logits according to its index in the vocabulary:
\begin{equation}
p(t_k) = {\rm softmax}\big({\rm LLM} (x;t_{history})\big)[t_k].
\end{equation}
The information gain defined above represents the uncertainty reduction of the masked part to the expected output.
For example, if the information gain is similar between complete and masked instructions, it indicates that the LLM may half-listen to the surface-pattern of the input instruction.

At each time step $t$, we randomly sample $N$ instances and compute their IG for each seen task.
Then we select $M$ seen tasks with the lowest mean IG as replay tasks, and merge the $|MN|$ instances into current training tasks $\mathcal{T}^t$.
Since our goal is to widen the gap between the complete (without mask) instance $d_\tau$ and the surface-level (with mask) instance $d_\tau^m$, the loss function is defined as:

\begin{equation}
\mathcal{L} = \mathcal{L}_{CE}(d_\tau) + \lambda\mathcal{L}_{JSD}\Big(\frac{p_t(d_\tau^m)}{\beta} \Big\Vert \frac{p_{t-1}(d_\tau^m)}{\beta}\Big),
\label{loss}
\end{equation}
where $\mathcal{L}_{CE}$ is cross entropy loss to maximize the ground truths.
$\mathcal{L}_{JSD}$ is Jensen–Shannon \cite{jsd} divergence of two distributions output by current LLM and frozen LLM from $t-1$, which is usually utilized as penalty in CL methods \cite{pcll,dcl,dynainst} to preserve original abilities.
However, we only apply JSD on masked instance $d_\tau^m$ rather than complete instance $d_\tau$.
And JSD value is symmetric and in $[0, 1]$ to easily balance $\mathcal{L}_{JSD}$ and $\mathcal{L}_{CE}$.
$\lambda$ is a hyperparameter that controls the weight of $\mathcal{L}_{JSD}$.
Moreover, $\beta$ is the dynamic temperature to soften probability distribution and is calculated as follows:
\begin{equation}
\beta = 2 - {\rm min}\big(\mathcal{G}(d_\tau, d_\tau^m), 1\big),
\end{equation}
where ${\rm min}$ represents the scaling of $\mathcal{G}(d_\tau, d_\tau^m)$ into the range $(-\infty, 1]$.
The lower the information gain, the greater the opportunity we give other tokens to improve the generalization ability of the LLM.
By doing so, $\mathcal{L}_{CE}$ maximizes likelihood for complete instances, and $\mathcal{L}_{JSD}$ dynamically adjusts the degree of conservatism when instructions are masked, enabling the LLM to be sensitive to key parts and alleviate the half-listening problem.
The detailed implementation is shown in Appendix~\ref{algorithm}.

\subsection{Evaluation Protocol}\label{3.4}
Since CIT trains tasks sequentially, we first introduce our construction method of multi-step datasets in this section.
Furthermore, different from using ROUGE-L \cite{rouge} as metric in previous methods \cite{dynainst, citb}, 
we propose a multi-dimensional evaluation method that pays more attention to the instruction-following ability of LLMs.

\paragraph{Data restructuring.}
We evaluate our method on SupNatInst and Domain datasets, where each task contains a task definition, a few demonstrations, and several instances.
SupNatInst consists of over 1000 NLP tasks and 76 categories (e.g., text classification,
information extraction and etc.) \cite{supernatural}.
We select 128 tasks in 40 categories from SupNatInst, 88 tasks are used for training $\mathcal{T}_{seen}$ and 40 as held-out tasks $\mathcal{T}_{unseen}$.
And our Chinese domain dataset has 20 tasks and 12 categories, where 13 tasks are used for training $\mathcal{T}_{seen}$ and 7 as held-out tasks $\mathcal{T}_{unseen}$.
We use two strategies, single-task (ST) and single-category (SC), to build multi-step training datasets.
For the ST setting, we fix $n$ equal to 1, where only 1 task in $\mathcal{T}_{seen}^t$ at time step $t$.
For the SC setting, we divide seen tasks into multiple groups according to their categories, and train different categories at each time step, because real training scenarios may gradually enhance model abilities of specific categories when training tasks are not available synchronously.
Furthermore, to enhance the balance and diversity of each test dataset while accelerating the evaluation process, we sample a few instances for each $\tau_{test}$ based on Self-BLEU \cite{selfbleu} score and label distribution.
More details about datasets can be found in Appendix~\ref{dataset}.

\begin{table*}
\setlength{\tabcolsep}{0.5mm}
\resizebox{\linewidth}{!}{%
\begin{tabular}{@{} l @{~} rr rr rr rr rr rr rr rr @{}}
    \toprule
		&\multicolumn{4}{c}{\textbf{Sup-NatInst-ST}}&\multicolumn{4}{c}{\textbf{Sup-NatInst-SC}}&\multicolumn{4}{c}{\textbf{Domain-ST}}&\multicolumn{4}{c}{\textbf{Domain-SC}}\\
	\cmidrule(r){2-5} \cmidrule(r){6-9} \cmidrule(r){10-13} \cmidrule(r){14-17}
   &\multicolumn{2}{c}{\textbf{Seen Tasks}}&\multicolumn{2}{c}{\textbf{Held-out Tasks}}&\multicolumn{2}{c}{\textbf{Seen Tasks}}&\multicolumn{2}{c}{\textbf{Held-out Tasks}}&\multicolumn{2}{c}{\textbf{Seen Tasks}}&\multicolumn{2}{c}{\textbf{Held-out Tasks}}&\multicolumn{2}{c}{\textbf{Seen Tasks}}&\multicolumn{2}{c}{\textbf{Held-out Tasks}}\\
	\cmidrule(r){2-3}\cmidrule(r){4-5}\cmidrule(r){6-7}\cmidrule(r){8-9}\cmidrule(r){10-11}\cmidrule(r){12-13}\cmidrule(r){14-15}\cmidrule(r){16-17}
    \textbf{Model} &\texttt{P-score} &\texttt{V-score} &\texttt{P-score} &\texttt{V-score} &\texttt{P-score} &\texttt{V-score} &\texttt{P-score} &\texttt{V-score} &\texttt{P-score} &\texttt{V-score} &\texttt{P-score} &\texttt{V-score} &\texttt{P-score} &\texttt{V-score} &\texttt{P-score} &\texttt{V-score}\\
    \midrule
    SFT & 35.1 & 12.0 & 25.9 & 24.1 & 51.1 & 4.5 & 34.2 & 6.7 & 43.5 & 12.0 & 37.0 & 16.3 & 52.2 & 8.3 & 43.1 & 10.5 \\
    LoRA & 33.7 & 12.4 & 26.7 & 23.0 & 48.7 & 4.7 & 36.1 & 5.3 & 41.8 & 12.8 & 38.2 & 15.9 & 49.5 & 8.9 & 44.6 & 10.0 \\
    L2 & 34.7 & 12.3 & 26.5 & 23.2 & 50.4 & 4.8 & 35.4 & 5.6 & 42.9 & 12.6 & 37.7 & 16.7 & 50.2 & 8.6 & 42.9 & 10.4 \\
    EWC & 30.2 & 13.5 & 25.1 & 24.6 & 47.9 & 5.9 & 33.6 & 7.4 & 41.4 & 13.2 & 35.8 & 17.9 & 48.7 & 10.4 & 41.5 & 11.8 \\
    \midrule
    DARE & - & - & - & - & 54.4 & 3.9 & 39.8 & 4.4 & - & - & - & - & 56.6 & \underline{5.7} & 45.9 & 10.1\\
    LM-Cocktail & - & - & - & - & 55.0 & 3.7 & 40.0 & 4.1 & - & - & - & - & 56.9 & 6.3 & 46.4 & 10.5\\
    \midrule
    PCLL & 50.5 & 5.4 & 38.2 & 5.6 & - & - & - & - & 52.4 & 10.8 & 43.7 & 14.6 & - & - & - & - \\
    DCL & 50.2 & 4.9 & 38.8 & 5.2 & - & - & - & - & 52.5 & 10.3 & 44.1 & 12.2 & - & - & - & - \\
    DYNAINST & 50.9 & 4.6 & 38.7 & 4.4 & 54.2 & 4.2 & 40.7 & 3.3 & 53.2 & \underline{9.1} & 44.6 & 10.9 & 56.3 & 8.3 & 47.2 & \underline{9.6} \\
    InsCL & \underline{52.5} & 4.0 & 38.4 & 5.5 & \underline{57.1} & 2.8 & 40.2 & 4.9 & - & - & - & - & - & - & - & - \\
    \midrule
    \textbf{KPIG} & 52.2 & \underline{3.5} & \underline{42.5} & \underline{1.7} & 56.5 & \underline{2.4} & \underline{43.6} & $\ast$1.2 & \underline{54.1} & $\ast$4.8 & \underline{47.8} & $\ast$3.3 & \underline{57.5} & $\ast$4.0 & \underline{49.7} & $\ast$2.7\\
    \midrule
    INIT &43.2 &5.3 & $\ast$43.8 & $\ast$1.5 & 43.2 &5.3 & $\ast$43.8 & \underline{1.5} & 28.9 & 10.8 & 39.1 & 13.5 & 28.9 & 10.8 & 39.1 & 13.5 \\
    MULTI & $\ast$59.8 & $\ast$2.2 & 41.4 & 4.2 & $\ast$59.8 & $\ast$2.2 & 41.4 & 4.2 & $\ast$60.0 & 9.7 & $\ast$49.9 & \underline{10.8} & $\ast$60.0 & 9.7 & $\ast$49.9 & 10.8 \\
    \bottomrule
    \end{tabular}}
    \caption{Performance of different methods on Sup-NatInst and Domain datasets. $\ast$ indicates the best, and $\_$ indicates the second best. The higher the \texttt{P-score}, the better the model performance. The lower the \texttt{V-score}, the stronger the instruction-following ability. Since \textbf{INIT} \textit{sometimes} serves as an upper bound for held-out tasks, and \textbf{MULTI} is \textit{sometimes} the upper bound for seen tasks, we report their results.}
    \label{Tab:main}
\end{table*}

\paragraph{Evaluation metrics.}
Previous methods use the ROUGE-L score to measure model performance \cite{dynainst, citb}, which may not comprehensively evaluate the instruction-following ability.
For example, \texttt{\{[1, 2, 3]\}} and \texttt{[1, 2, 3]} have same Rouge-1 scores with the ground truth \texttt{[1, 2]}, but the instruction explicitly requires generating a one-dimensional list format.
We evaluate model on $\tau_{test}$ of trained (seen) task set $\mathcal{T}_{seen}$ and held-out (unseen) task set $\mathcal{T}_{unseen}$ with the following metrics:
\begin{itemize}[itemsep=2pt, topsep=0pt, parsep=0pt]
\item \textbf{WFR} measures the wrong-format rate of tasks that instructions in them explicitly constrain delimiters, sequence, formats, or length limits.
\item \textbf{OOS} measures the out-of-scope rate of classification or extraction tasks whose instructions constrain output choices.
\item \textbf{WR} measures wordy rate when the length of responses are greater than the threshold.
\item \textbf{F1} measures the performance for sequence labeling tasks.
\item \textbf{ACC} measures the precision for classification tasks or execution accuracy for code tasks.
\item \textbf{ROUGE} and \textbf{BLEU} measures the similarity for tasks such as summarization.
\item \textbf{Match} measures the match rate for tasks that the ground truths are unordered sets.
\item \textbf{GPT} leverages GPT-4 to measure whether tasks of generating open-ended short texts are reasonable, which require commonsense or reasoning skills to verify.
\end{itemize}
Then we use \texttt{P-score} and \texttt{V-score} to measure the performance and instruction-following ability.
\texttt{P-score} is the average of F1, ACC, ROUGE, BLEU, Match, and GPT, which can measure generalization ability on held-out tasks.
\texttt{V-score} is the average of WFR, OOS, and WR, acting as an indicator of instruction-violation degree.

\section{Experiments}

\subsection{Experimental Setup}

\paragraph{Baselines.}
We compare our method in the CIT setting with the following baselines.
\textbf{INIT} is the foundation LLM without training.
\textbf{MULTI} shuffles instances in all training tasks and trains them together.
\textbf{SFT} \cite{instructgpt} directly fine tunes the LLM on seen tasks sequentially.
\textbf{LoRA} \cite{lora} updates the low-rank matrices while the LLM backbone is fixed.
\textbf{L2} and \textbf{EWC} \cite{ewc} mitigate forgetting by regularizing the loss to penalize the changes of important parameters.
\textbf{DARE} \cite{dare} and \textbf{LM-Cocktail} \cite{lmcocktail} obtain the target LLM 
by model merging, which train multiple models on different tasks and merge them into a single model through weighted average.
\textbf{DYNAINST} \cite{dynainst} and \textbf{InsCL} \cite{InsCL} dynamically determines which instances are stored and replayed based on their proposed metrics.
\textbf{PCLL} \cite{pcll} and \textbf{DCL} \cite{dcl} generate pseudo samples for history tasks and utilize knowledge distillation strategy to mitigate catastrophic forgetting.

\paragraph{Hyperparameters.}
We choose LLaMA-2-7B-Chat \cite{llama2} and baichuan-vicuna-chinese-7b \footnote{\url{https://huggingface.co/fireballoon/baichuan-vicuna-chinese-7b}} as foundation models for experiments on SupNatInst and Domain datasets respectively.
Our experiments are implemented based on DeepSpeed \cite{deepspeed} and FastChat \cite{fastchat}. 
And $8$ NVIDIA A100 GPUs are used.
We optimize the model parameters by using AdamW optimizer~\citep{adamw} with the learning rate of $2e-5$.
The batch size is $384$ with $16$ gradient accumulation steps and $3$ sentences per GPU.
We conduct a grid search to find other hyperparameters that maximize the average \texttt{P-score} on seen and held-out tasks.
The optimal settings are: $\{\alpha=0.3, \lambda=0.02, M=10, N=10, epoch=1\}$ on Sup-NatInst and $\{\alpha=0.6, \lambda=0.01, M=3, N=100, epoch=1\}$ on Domain.
Additionally, we iteratively perform $30$ evolutions for each task in the instruction-diversity stage. 
And when evaluating held-out tasks, we add $2$ additional pre-written demonstrations to the input context.
More detailed information about the implementation can be found in Appendix~\ref{algorithm}.

\subsection{Main Results}

Table~\ref{Tab:main} summarizes the performances of different methods.
It should be noted that \textbf{INIT} is a pretrained LM, and \textbf{MULTI} trains the LLM with all seen tasks together, so they have the same results on ST and SC. 
We train \textbf{PCLL} and \textbf{DCL} only on the datasets constructed by the ST strategy which are designed to learn single-task parameters sequentially.
Moreover, we only conduct SC experiments on \textbf{DARE} and \textbf{LM-Cocktail} which merge peer models on each category, because training a sub-LLM for each task requires a much larger resource than training a sub-LLM for each category.
Our observations are summarized as follows.

Firstly, foundation LLMs require more training on domain-specific datasets to achieve performance improvements.
In the benchmark of CIT \cite{citb}, multi-task learning (\textbf{MULTI}) is served as an upper bound on seen tasks while \textbf{INIT} is the upper bound for held-out tasks.
The difference is that \textbf{MULTI} achieves the best \texttt{P-score} on held-out tasks of Domain.
This may be because our domain-specific dataset is highly specialized, which leaves the foundational model (\textbf{INIT}) lacking in pertinent knowledge without training.
In addition, compared with the held-out results of \textbf{INIT} on Sup-NatInst, most methods show performance degradation of \texttt{P-score} and \texttt{V-score}, which may indicate forgetting ability in the foundation LLM.

Secondly, the catastrophic forgetting problem of the single-task setting is more severe than the single-category setting.
The performance of \textbf{SFT}, \textbf{LoRA}, \textbf{L2}, and \textbf{EWC} on seen tasks and held-out tasks under the ST setting is significantly worse than the SC setting, while the performance gap between ST and SC on \textbf{DYNAINST} and \textbf{KPIG} is relatively small.
Furthermore, \textbf{MULTI} stands out with the highest \texttt{P-score} on all seen tasks. 
The above phenomenons indicate that the training difficulty and overfitting become more pronounced when training on a single task sequentially, and mixing data from different tasks and replaying data can help mitigate performance degradation.

Thirdly, model-merge methods perform better on seen tasks than on held-out tasks. This may be because they selectively inherit abilities of different task categories from multiple models, but abilities are limited when faced with held-out tasks.

\begin{table}
\centering
\small
\resizebox{\linewidth}{!}{
\begin{tabular}{lrrrr}
\toprule
& \multicolumn{2}{c}{\textbf{Seen Tasks}} & \multicolumn{2}{c}{\textbf{Held-out Tasks}} \\
\cmidrule(r){2-3}\cmidrule(r){4-5}
\textbf{Model} & \texttt{P-score} & \texttt{V-score} & \texttt{P-score} & \texttt{V-score}\\
\midrule
w/o div & \underline{52.5} & \underline{4.1} & \underline{41.6} & \underline{3.2}\\
w/o mask & 48.3 & 4.9 & 41.2 & 3.6\\
w/o jsd & $\ast$52.9 & 4.7 & 39.4 & 5.9\\
w/o temp & 51.7 & 4.4 & 40.9 & 4.0\\
\midrule
\textbf{KPIG} & 52.2 & $\ast$3.5 & $\ast$42.5 & $\ast$1.7\\
\bottomrule
\end{tabular}}
\caption{Ablation studies on Sup-NatInst-ST.}
\label{Tab:ablation}
\end{table} 

Finally, our proposed \textbf{KPIG} achieves the best performance, especially on held-out tasks and the instruction-following ability (\texttt{V-score}).
The \texttt{P-score} of \textbf{KPIG} on seen tasks is slightly lower than the \textbf{InsCL}, which may be related to the replayed number.
On the held-out tasks, our method performs significantly better than other CL baselines in both \texttt{P-score} and \texttt{V-score}, which shows stronger generalization ability and instruction-following ability.
Moreover, the \texttt{V-score} of other baselines on the Domain dataset is much larger than the Sup-NatInst dataset, while our method maintains lower \texttt{V-score} on both Sup-NatInst and domain-specific datasets.
This indicates half-listening and instruction violations may be more likely to occur on a specific domain, and our method can better capture the task-aware information and improve the instruction-following ability.

\subsection{Ablation Study}

To evaluate the effectiveness of each component in \textbf{KPIG}, we conduct ablation studies on the Sup-NatInst-ST dataset. 
Firstly, we remove the instruction diversity module (\textbf{w/o div}) and only extract key parts for the initial instruction of each task. 
Then, to investigate the significance of our key-part mask mechanism, we remove the mask step (\textbf{w/o mask}).
The \textbf{w/o mask} setting replays data based on predictive entropy like \textbf{DYNAINST} and performs JSD on the predictive distribution of the complete instruction of current LLM and frozen LLM.
Finally, we investigate the effects of removing $\mathcal{L}_{JSD}$ (\textbf{w/o jsd}) and dynamic temperature (\textbf{w/o temp}).

The results are shown in Table~\ref{Tab:ablation}. 
When only initial instructions for each task are used without diversification (\textbf{w/o div}), the \texttt{P-score} of seen tasks is slightly higher than \textbf{KPIG}, but the performance of held-out tasks become worse, indicating that increasing data diversity helps alleviate overfitting and preserve generalization.
In the \textbf{w/o mask} setting, the \texttt{V-score} drops significantly and the \texttt{P-score} of seen tasks is much lower than \textbf{KPIG}.
It proves the effectiveness of measuring and learning task-aware information by masking key-part in instructions, which assists LLMs in comprehending the tasks to be executed rather than simply maintaining the original ability.
The results of \textbf{w/o jsd} and \textbf{w/o temp} on held-out tasks suggest that they are helpful in maintaining instruction-following and generalization abilities. 
Moreover, the decline in \textbf{w/o mask} results for seen tasks and in \textbf{w/o jsd} for held-out tasks suggests an interdependence between key-part mask and $\mathcal{L}_{JSD}$.
Without $\mathcal{L}_{JSD}$ and key-part mask, LLMs may struggle to widen the gap between task-aware constraints in key parts and some general descriptions in instructions, which is crucial for balancing learning new information with maintaining original judgments.

\begin{figure}
    \centering
    \includegraphics[width=3.0in]{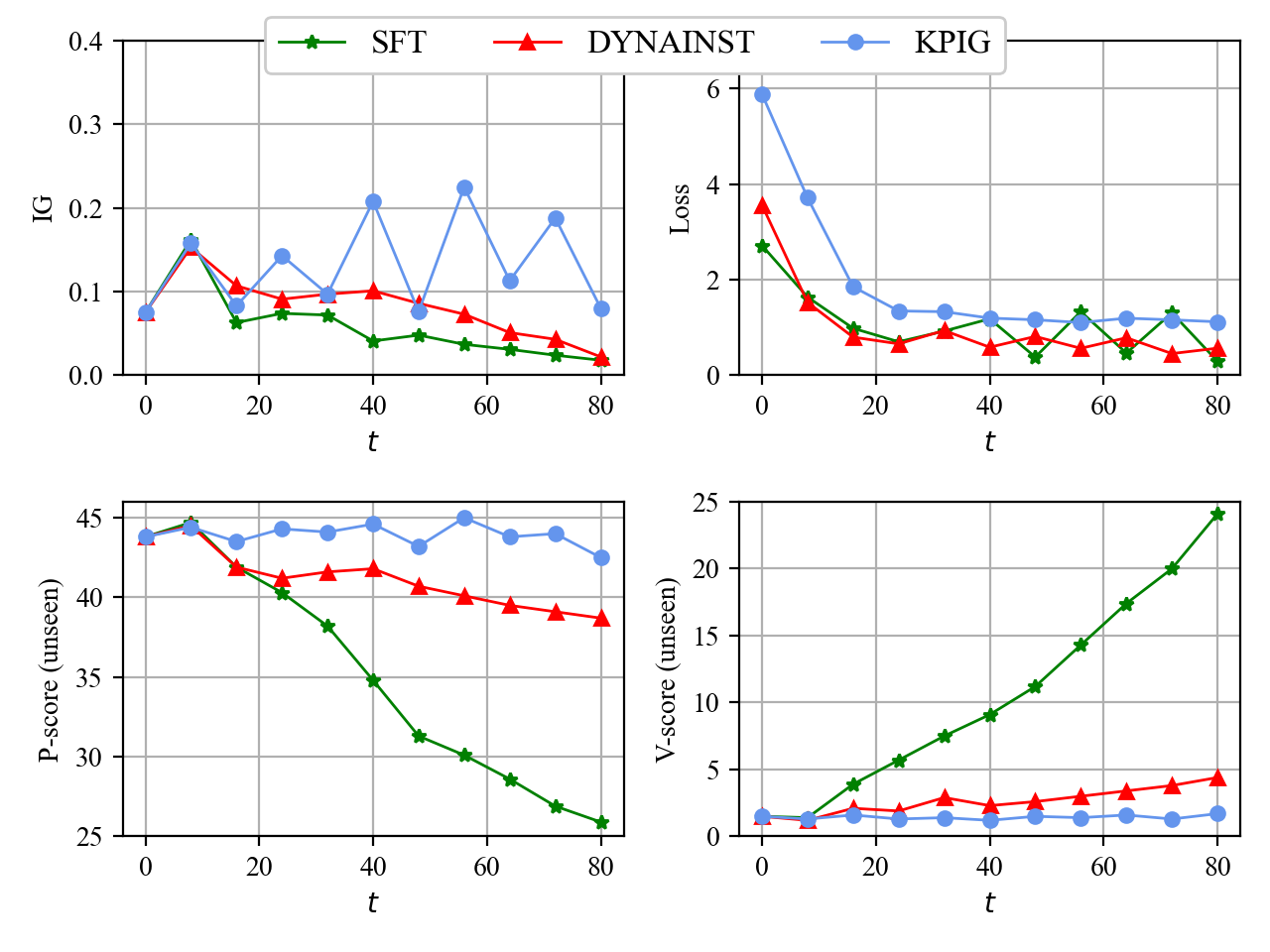}
    \caption{The changing trends of information gain, loss, \texttt{P-score}, and \texttt{V-score} on Sup-NatInst-ST over steps.
}
    \label{fig:info_gain}
\end{figure}

\subsection{Investigations on Information Gain}\label{Investigations on Information Gain}

Herein, we investigate the correlation between our information gain and the instruction-following ability on held-out tasks and held-out instructions.

\paragraph{Overfitting.} 
As shown in Figure~\ref{fig:info_gain}, the IG of our method oscillates above the initial value, while other methods begin to decline at approximately $t=20$. 
This interesting oscillation may be related to our replay mechanism based on IG, which chooses tasks with the lowest IG. 
The curve gradually rises when IG is low, and then falls back to the level of the foundation model when it reaches the upper bound.
Meanwhile, the changing trends of \texttt{P-score} and \texttt{V-score} of held-out tasks are in alignment with information gain, indicating the validity of employing IG as a metric for measuring task-aware ability. 
In addition, compared with other methods, our loss progression maintains a more stable and smooth decline.
This may be because our method can effectively alleviate overfitting on individual tasks and does not require more recalibrations after training previous tasks.

\begin{table}
\centering
\small
\resizebox{\linewidth}{!}{
\begin{tabular}{l|rrrrrr}
\toprule
\textbf{Metric} & Ins$_1$ & Ins$_2$ & Ins$_3$ & Ins$_4$ & Ins$_5$ & Ins$_6$\\
\midrule
P-score & 68.0 & 67.0 & 68.0 & 70.0 & 72.0 & 66.0\\
V-score & 4.0 & 14.0 & 8.0 & 3.0 & 3.0 & 15.0\\
IG & 0.28 & 0.22 & 0.26 & 0.29 & 0.30 & 0.15\\
\bottomrule
\end{tabular}}
\caption{Results of the smcalflow classification task (one of seen tasks) on 6 held-out instructions.}
\label{Tab:held_out}
\end{table} 

\begin{table}
\centering
\small
\resizebox{\linewidth}{!}{
\begin{tabular}{lrrrrrr}
\toprule
\textbf{Response} & INIT & MULTI & SFT & LM-Cocktail & DYNAINST & KPIG\\
\midrule
\texttt{\underline{U}ser}   & 0.0 & 49.0 & 0.0 & 4.0 & 85.0 & 14.0 \\
\texttt{\underline{A}gent}  & 0.0 & 50.0 & 0.0 & 58.0 & 15.0 & 8.0 \\
\texttt{\underline{u}ser} (required)   & 1.0 & 0.0 & 0.0 & 15.0 & 0.0 & 45.0 \\
\texttt{\underline{a}gent} (required)  & 69.0 & 0.0 & 0.0 & 9.0 & 0.0 & 26.0 \\
\midrule
IG                & 0.13 & 0.07 & 0.00 & 0.11 & 0.04 & 0.19 \\
\bottomrule
\end{tabular}}
\caption{Statistics of responses after modifying constraints in the smcalflow classification instruction. It should be noted that this task requires the first letter of \texttt{\underline{U}ser} and \texttt{\underline{A}gent} to be capitalized during training, and we require \texttt{\underline{u}ser} and \texttt{\underline{a}gent} during testing.}
\label{Tab:mislead}
\end{table} 

\paragraph{Held-out instructions.}
To further explore the instruction-following and generalization abilities, we modify the instruction of smcalflow classification task after training.
We collect 6 held-out instructions which are not seen during training.
As Table~\ref{Tab:held_out} shows, the \texttt{P-score} and \texttt{V-score} on held-out instructions are significantly correlated with information gain, indicating that information gain can be used to measure the generalization ability and instruction-following ability.

In addition, as shown in Table~\ref{Tab:mislead}, we modify the constraints of capital letters (\texttt{Answer with \underline{U}ser or \underline{A}gent}) to obtain the misleading constraints (\texttt{Answer with \underline{u}ser or \underline{a}gent}). 
Most of the responses of \textbf{INIT} are legal, indicating that the initial foundation LLM has strong instruction-following ability.
\textbf{LM-Cocktail} gives a small proportion of legal responses because model-merging methods can inherit abilities of other LLMs.
All responses of \textbf{MULTI} and \textbf{DYNAINST} are illegal, which means they are overfitting to training instructions and half-listen to the misleading instruction during testing.
The responses given by \textbf{SFT} are all irrelevant due to catastrophic forgetting in CL, which forgets not only historical tasks but also the ability of the foundation LLM.
Moreover, $71\%$ responses of our method are \texttt{\underline{u}ser} and \texttt{\underline{a}gent}, and our average information gain on the misleading instruction is the highest, which shows that \textbf{KPIG} has a stronger ability to alleviate the half-listening problem even if similar instructions are seen during training.

\section{Conclusion}

In this paper, we propose a novel CIT method to alleviate catastrophic forgetting and half-listening problems, which enables LLMs to be sensitive to task-specific constraints of both seen and held-out tasks.
Our method calculates the information gain of masked key parts, to selectively replay historical data and dynamically adjust the temperature.
Experimental results show strong instruction-following and generalization abilities in comparison to other continual learning methods.
Furthermore, our investigation into the proposed \texttt{P-score}, \texttt{V-score}, and IG not only confirms their relevance in model performance and instruction adherence, but also demonstrates that our method effectively alleviates overfitting to seen-task instruction and maintains the generalization ability.

\clearpage

\section*{Acknowledgments}

This work is supported by the fund of Laboratory for Advanced Computing and Intelligence Engineering.
We also appreciate the assistance from Hongbin Zhang, Xinyu Wang, and Chen Shi of Nanjing University of Science and Technology. 
We thank all anonymous reviewers for their valuable comments and suggestions on this work.

\section*{Limitations}

In this paper, we use GPT-4 to extract key parts of instructions and diversify instructions, but the gap between this method and manual writing in controllability and accuracy is not fully evaluated. 
We propose WFR, OOS, and WR as evaluation dimensions of the instruction-following ability based on manually annotating explicit constraints in instructions.
However, there may be other constraints or ways for evaluating the instruction-following ability that exist and deserve to be considered.
Moreover, we dynamically replay instances and adjust the training objective by calculating information gain of key parts, making the LLM more sensitive to task-specific constraints in instructions and thereby alleviating the half-listening problem. 
Our experiments (Table~\ref{Tab:mislead}) also find that such half-listening problem also occurs in multi-task learning, so the implications of our mask information gain on other natural language processing tasks involving LLMs and the effects of masking other parts (e.g., context, demonstrations) within instances can be explored in the future.

\section*{Ethics Statement}

In this paper, we propose a novel CIT paradigm to alleviate the half-listening problem, which aims to improve instruction-following ability and generalization ability of LLMs.
Our experiments are conducted with the publicly available Super-NaturalInstructions dataset, our in-house dataset, and LLMs from open sources, one of whose initial intentions is to promote the development of instruction-based LLMs.
Since LLMs trained with web data may produce toxic content, we must state that the texts generated by our method do not represent our opinions.
To alleviate such potential negative impacts, we can adopt appropriate detoxification strategies and principle constraints, and we encourage future work to explore these issues.

\bibliography{custom}

\newpage
\appendix

\begin{table*}
\centering
\small
\begin{tabular}{p{\textwidth}}
\toprule
\multicolumn{1}{c}{\textbf{Key-part extraction}} \\
\midrule
\midrule
\texttt{What is the \#key part\# in the \#instruction\#?}\\
\texttt{The \#key part\# refers to the consecutive span in the \#instruction\# that has guiding significance for the format, length, content, and rationality of the ground truth when bridging \#input\# to \#output\#.}\\
\texttt{Please return key parts as a list.}\\
\texttt{\#instruction\#:}\\
\texttt{\{...\}}\\
\texttt{\#input\#:}\\
\texttt{\{...\}}\\
\texttt{\#output\#:}\\
\texttt{\{...\}}\\
\midrule
\multicolumn{1}{c}{\textbf{\Square Concretizing \CheckedBox Reasoning \Square Constraint}} \\
\midrule
\midrule
\texttt{I want you act as an Instruction Creator.}\\
\texttt{Your goal is to draw inspiration from the \#Given Instruction\# and \#Key Part\# to create a brand new instruction \#Created Instruction\#.}\\
\texttt{The \#Created Instruction\# must be reasonable and must be understood and responded by humans.}\\
\texttt{And this \#Created Instruction\# can guide the \#Input\# to give the \#Output\#.}\\
\texttt{Your \#Created Instruction\# cannot omit the non-text parts such as the table and code in the \#Given Instruction\#.}\\
\texttt{You should complicate the \#Given Instruction\# using the following method:}\\
\Square\texttt{\textbf{Please replaces general concepts in \#Key Part\# with more specific concepts.}}\\
\CheckedBox\texttt{\textbf{If \#Key Part\# can be organized into a few simple thinking processes, you can rewrite it to explicitly request multiple-step reasoning.}}\\
\Square\texttt{\textbf{Please add one more constraints/requirements into \#Given Instruction\#.}}\\
\texttt{You should try your best not to make the \#Created Instruction\# become verbose, \#Created Instruction\# can only add 10 to 20 words into the \#Given Instruction\#.}\\
\texttt{'\#Given Instruction\#', '\#Created Instruction\#', 'given instruction' and 'created instruction' are not allowed to appear in \#Created Instruction\#.}\\
\texttt{\#Given Instruction\#:}\\ 
\texttt{\{...\}}\\
\texttt{\#Key Part\#:}\\ 
\texttt{\{...\}}\\
\texttt{\#Input\#:}\\ 
\texttt{\{...\}}\\
\texttt{\#Output\#:}\\ 
\texttt{\{...\}}\\
\texttt{\#Created Instruction\#:}\\
\midrule
\multicolumn{1}{c}{\textbf{Breadth}} \\
\midrule
\midrule
\texttt{I want you act as a Instruction Rewriter.}\\
\texttt{Your goal is to draw inspiration from the \#Given Instruction\# and \#Key Part\# to rewrite a brand new instruction \#Rewritten Instruction\#.}\\
\texttt{This \#Rewritten Instruction\# should belong to the same domain as the \#Given Instruction\# but be even more rare.}\\
\texttt{And this \#Rewritten Instruction\# can guide the \#Input\# to give the \#Output\#.}\\
\texttt{\textbf{\#Key Part\# in the \#Given Instruction\# should be unchanged.}}\\
\texttt{The LENGTH and complexity of the \#Rewritten Instruction\# should be similar to that of the \#Given Instruction\#.}\\
\texttt{The \#Rewritten Instruction\# must be reasonable and must be understood and responded by humans.}\\
\texttt{'\#Given Instruction\#', '\#Rewritten Instruction\#', 'given instruction' and 'rewritten instruction' are not allowed to appear in \#Rewritten Instruction\#.}\\
\texttt{\#Given Instruction\#:}\\ 
\texttt{\{...\}}\\
\texttt{\#Key Part\#:}\\ 
\texttt{\{...\}}\\
\texttt{\#Input\#:}\\ 
\texttt{\{...\}} \\
\texttt{\#Output\#:}\\ 
\texttt{\{...\}}\\
\texttt{\#Rewritten Instruction\#:}\\
\bottomrule
\end{tabular}
\caption{Our templates for extracting key parts and evolving instructions.}
\label{Tab:template}
\end{table*}

\begin{table}
	\resizebox{\linewidth}{!}{
    \begin{tabular}{l|ccccc|c}  
    \toprule
	\textbf{Dataset} & Task & Category & Training & Test & Held-out & Time Step\\  
    \midrule
	Sup-NatInst-ST & 128 & 40 & 50,901 & 12,800 & 40 & 88\\
	Sup-NatInst-SC & 128 & 40 & 50,901 & 12,800 & 40 & 34 \\
	Domain-ST & 20 & 12 & 52,000 & 10,000 & 7 & 13\\
	Domain-SC & 20 & 12 & 52,000 & 10,000 & 7 & 9\\
    \bottomrule
    \end{tabular}}
    \caption{Statistics of datasets.}
\label{Statistics of datasets}
\end{table}

\begin{table*}
\centering
\scriptsize
\begin{tabular}{l|ccccc}
\toprule
\textbf{Name} & \textbf{Category} & \textbf{Metric} & \textbf{Scope} & \textbf{Format} & \textbf{Usage} \\
\midrule
mctaco\_wrong\_answer\_generation\_event\_ordering & Wrong Candidate Generation & GPT & - & - & train \\
\midrule
mctaco\_grammatical\_logical & Text Quality Evaluation & ACC & In choice & - & train \\
\midrule
essential\_terms\_identifying\_essential\_words & Question Understanding & F1 & - & Split by , & train \\
\midrule
multirc\_classify\_correct\_answer & Answer Verification & ACC & In choice & - & train \\
\midrule
squad11\_question\_generation & Question Generation & ROUGE & - & - & train \\
\midrule
conala\_remove\_duplicates & Program Execution & Match & - & List & train \\
\midrule
commongen\_sentence\_generation & Data to Text & ROUGE & - & - & train \\
\midrule
story\_cloze-rocstories\_sentence\_generation & Text Completion & ROUGE & - & - & train \\
\midrule
zest\_text\_modification & Question Rewriting & ROUGE & - & - & train \\
\midrule
detoxifying-lms\_classification\_fluency & Text Completion & ACC & In choice & - & train \\
\midrule
afs\_argument\_quality\_death\_penalty & Text Matching & ACC & In choice & - & train \\
\midrule
count\_nouns\_verbs & Pos Tagging & ACC & - & Number & train \\
\midrule
snli\_contradiction\_to\_entailment\_text\_modification & Sentence Composition & ROUGE & - & - & train \\
\midrule
snli\_classification & Textual Entailment & ACC & In choice & - & train \\
\midrule
hotpotqa\_sentence\_generation & Explanation & ROUGE & - & - & train \\
\midrule
iirc\_link\_exists\_classification & Answerability Classification & ACC & In choice & - & train \\
\midrule
stereoset\_sentence\_generation\_antistereotype & Fill in The Blank & GPT & - & - & train \\
\midrule
dream\_incorrect\_answer\_generation & Wrong Candidate Generation & ROUGE & - & - & train \\
\midrule
tellmewhy\_question\_answerability & Answerability Classification & ACC & In choice & - & train \\
\midrule
(296)storycloze\_correct\_end\_classification & Text Completion & ACC & In choice & - & train \\
\midrule
(298)storycloze\_correct\_end\_classification & Coherence Classification & ACC & In choice & - & train \\
\midrule
numeric\_fused\_head\_resolution & Coreference Resolution & ACC & In choice & - & train \\
\midrule
stereoset\_classification\_profession & Stereotype Detection & ACC & In choice & - & train \\
\midrule
jigsaw\_classification\_obscene & Toxic Language Detection & ACC & In choice & - & train \\
\midrule
winomt\_classification\_gender\_anti & Gender Classification & ACC & In choice & - & train \\
\midrule
winomt\_classification\_profession\_pro & Gender Classification & ACC & In choice & - & train \\
\midrule
squad20\_answerable\_unanswerable\_question\_classification & Answerability Classification & ACC & In choice & - & train \\
\midrule
winomt\_classification\_gender\_identifiability\_anti & Gender Classification & ACC & In choice & - & train \\
\midrule
casino\_classification\_negotiation\_vouch\_fair & Negotiation Strategy Detection & ACC & In choice & - & train \\
\midrule
inverse\_causal\_relationship & Cause Effect Classification & ACC & In choice & - & train \\
\midrule
numeric\_fused\_head\_reference & Coreference Resolution & ACC & In context & - & train \\
\midrule
com\_qa\_paraphrase\_question\_generation & Question Rewriting & ROUGE & - & - & train \\
\midrule
scruples\_anecdotes\_title\_generation & Title Generation & ROUGE & - & - & train \\
\midrule
senteval\_odd\_word\_out & Linguistic Probing & ACC & In choice & - & train \\
\midrule
aquamuse\_answer\_given\_in\_passage & Answerability Classification & ACC & In choice & - & train \\
\midrule
udeps\_eng\_coarse\_pos\_tagging & Pos Tagging & ACC & In choice & - & train \\
\midrule
multi\_woz\_classification & Speaker Identification & ACC & In choice & - & train \\
\midrule
esnli\_classification & Textual Entailment & ACC & In choice & - & train \\
\midrule
extreme\_abstract\_summarization & Summarization & ROUGE & - & - & train \\
\bottomrule
\end{tabular}
\caption{Details of 1-40 task in the SupNatInst dataset.}
\label{Tab:SupNatInst40}
\end{table*}

\begin{table*}
\centering
\scriptsize
\begin{tabular}{l|ccccc}
\toprule
\textbf{Name} & \textbf{Category} & \textbf{Metric} & \textbf{Scope} & \textbf{Format} & \textbf{Usage} \\
\midrule
ambigqa\_text\_generation & Question Rewriting & ROUGE & - & - & train \\
\midrule
mmmlu\_answer\_generation\_computer\_security & Question Answering & ACC & In choice & - & train \\
\midrule
mmmlu\_answer\_generation\_world\_religions & Question Answering & ACC & In choice & - & train \\
\midrule
protoqa\_question\_generation & Question Generation & ROUGE & - & - & train \\
\midrule
copa\_commonsense\_reasoning & Cause Effect Classification & ACC & In choice & - & train \\
\midrule
copa\_commonsense\_cause\_effect & Cause Effect Classification & ACC & In choice & - & train \\
\midrule
synthetic\_multiply\_evens & Program Execution & Match & - & List & train \\
\midrule
synthetic\_multiply\_odds & Program Execution & Match & - & List & train \\
\midrule
cfq\_mcd1\_explanation\_to\_sql & Text to Code & GPT & - & - & train \\
\midrule
cfq\_mcd1\_sql\_to\_explanation & Text to Code & ACC & In choice & - & train \\
\midrule
freebase\_qa\_topic\_generation & Question Understanding & ROUGE & - & - & train \\
\midrule
dialogre\_identify\_names & Speaker Identification & ACC & - & - & train \\
\midrule
coached\_conv\_pref\_classifier & Speaker Identification & ACC & In choice & - & train \\
\midrule
defeasible\_nli\_atomic\_classification & Textual Entailment & ACC & In choice & - & train \\
\midrule
librispeech\_asr\_text\_auto\_completion & Text Completion & ROUGE & - & - & train \\
\midrule
librispeech\_asr\_missing\_word\_prediction & Fill in The Blank & GPT & - & - & train \\
\midrule
bard\_analogical\_reasoning\_affordance & Word Analogy & GPT & Noun & - & train \\
\midrule
bard\_analogical\_reasoning\_travel & Word Analogy & GPT & - & - & train \\
\midrule
bard\_analogical\_reasoning\_trash\_or\_treasure & Word Analogy & GPT & - & - & train \\
\midrule
penn\_treebank\_coarse\_pos\_tagging & Pos Tagging & ACC & In choice & - & train \\
\midrule
atomic\_classification\_causes & Commonsense Classification & ACC & In choice & - & train \\
\midrule
hrngo\_quality\_classification & Text Quality Evaluation & ACC & In choice & - & train \\
\midrule
glue\_mrpc\_paraphrasing & Text Matching & ACC & In choice & - & train \\
\midrule
wiki\_qa\_answer\_verification & Answer Verification & ACC & In choice & - & train \\
\midrule
amazonreview\_summary\_classification & Summarization & ACC & In choice & - & train \\
\midrule
numer\_sense\_multiple\_choice\_qa\_generation & Fill in The Blank & ACC & In choice & - & train \\
\midrule
cb\_entailment & Textual Entailment & ACC & In choice & - & train \\
\midrule
wscfixed\_coreference & Coreference Resolution & ACC & In choice & - & train \\
\midrule
dart\_question\_generation & Data to Text & ROUGE & - & Contain \_ & train \\
\midrule
gene\_extraction\_chemprot\_dataset & Named Entity Recognition & F1 & In context & - & train \\
\midrule
chemical\_extraction\_chemprot\_dataset & Named Entity Recognition & F1 & - & One answer & train \\
\midrule
hatexplain\_classification & Toxic Language Detection & ACC & In choice & - & train \\
\midrule
imppres\_longtextgeneration & Sentence Composition & GPT & - & - & train \\
\midrule
daily\_dialog\_question\_classification & Dialogue Act Recognition & ACC & In choice & - & train \\
\midrule
parsed\_pdfs\_summarization & Title Generation & ROUGE & - & - & train \\
\midrule
scitail\_classification & Textual Entailment & ACC & In choice & - & train \\
\midrule
blimp\_binary\_classification & Linguistic Probing & ACC & In choice & - & train \\
\midrule
bless\_hypernym\_generation & Word Semantics & ROUGE & - & - & train \\
\midrule
scifact\_title\_generation & Title Generation & ROUGE & - & - & train \\
\bottomrule
\end{tabular}
\caption{Details of 41-80 task in the SupNatInst dataset.}
\label{Tab:SupNatInst80}
\end{table*}

\begin{table*}
\centering
\scriptsize
\begin{tabular}{l|ccccc}
\toprule
\textbf{Name} & \textbf{Category} & \textbf{Metric} & \textbf{Scope} & \textbf{Format} & \textbf{Usage} \\
\midrule
smcalflow\_classification & Speaker Identification & ACC & In choice & - & train \\
\midrule
disfl\_qa\_text\_modication & Question Rewriting & GPT & - & - & train \\
\midrule
medical\_question\_pair\_dataset\_text\_classification & Text Matching & ACC & In choice & - & train \\
\midrule
winobias\_text\_generation & Coreference Resolution & Match & In context & Split by , & train \\
\midrule
civil\_comments\_threat\_classification & Toxic Language Detection & ACC & In choice & - & train \\
\midrule
civil\_comments\_insult\_classification & Toxic Language Detection & ACC & In choice & - & train \\
\midrule
web\_nlg\_data\_to\_text & Data to Text & GPT & - & - & train \\
\midrule
quartz\_question\_answering & Question Answering & ACC & In context & - & train \\
\midrule
mctaco\_wrong\_answer\_generation\_absolute\_timepoint & Wrong Candidate Generation & GPT & - & - & test \\
\midrule
mctaco\_span\_based\_question & Answerability Classification & ACC & In choice & - & test \\
\midrule
winogrande\_question\_generation\_person & Question Generation & GPT & - & - & test \\
\midrule
ropes\_story\_generation & Story Composition & ROUGE & - & - & test \\
\midrule
abductivenli\_classification & Coherence Classification & ACC & In choice & - & test \\
\midrule
scan\_structured\_text\_generation\_command\_action\_short & Text to Code & Match & In choice & Split by \_ & test \\
\midrule
odd-man-out\_classification\_no\_category & Word Semantics & ACC & In context & - & test \\
\midrule
combinations\_of\_list & Program Execution & Match & - & 2D list & test \\
\midrule
rocstories\_correct\_ending\_classification & Text Completion & ACC & In choice & - & test \\
\midrule
rocstories\_title\_answer\_generation & Title Generation & ROUGE & - & Length <= 3 & test \\
\midrule
dream\_classification & Question Understanding & ACC & In choice & - & test \\
\midrule
scruples\_event\_time & Text Categorization & ACC & In choice & - & test \\
\midrule
stereoset\_classification\_race & Stereotype Detection & ACC & In choice & - & test \\
\midrule
gap\_answer\_generation & Coreference Resolution & ACC & - & - & test \\
\midrule
winomt\_classification\_gender\_pro & Gender Classification & ACC & In choice & - & test \\
\midrule
hybridqa\_answer\_generation & Pos Tagging & ACC & In choice & - & test \\
\midrule
casino\_classification\_negotiation\_small\_talk & Negotiation Strategy Detection & ACC & In choice & - & test \\
\midrule
grailqa\_paraphrase\_generation & Question Rewriting & ROUGE & - & - & test \\
\midrule
persent\_sentence\_sentiment\_verification & Sentiment Analysis & ACC & In choice & - & test \\
\midrule
senteval\_inversion & Linguistic Probing & ACC & In choice & - & test \\
\midrule
mwsc\_question\_generation & Question Generation & ROUGE & - & - & test \\
\midrule
scruples\_anecdotes\_whoiswrong\_classification & Ethics Classification & ACC & In choice & - & test \\
\midrule
argument\_consequence\_classification & Text Matching & ACC & In choice & - & test \\
\midrule
glucose\_cause\_event\_detection & Cause Effect Classification & GPT & - & - & test \\
\midrule
google\_wellformed\_query\_sentence\_generation & Text Quality Evaluation & ACC & In context & - & test \\
\midrule
mmmlu\_answer\_generation\_international\_law & Question Answering & ACC & In choice & - & test \\
\midrule
glucose\_reverse\_cause\_emotion\_detection & Information Extraction & ROUGE & - & A >Causes> B & test \\
\midrule
conv\_ai\_2\_classification & Speaker Identification & ACC & In choice & - & test \\
\midrule
gap\_fill\_the\_blank\_coreference\_resolution & Coreference Resolution & ACC & In choice & - & test \\
\midrule
defeasible\_nli\_snli\_classification & Textual Entailment & ACC & In choice & - & test \\
\midrule
bard\_analogical\_reasoning\_causation & Word Analogy & GPT & - & - & test \\
\midrule
atomic\_classification\_xneed & Commonsense Classification & ACC & In choice & - & test \\
\midrule
atomic\_answer\_generation & Fill in The Blank & GPT & - & One answer & test \\
\midrule
superglue\_multirc\_answer\_verification & Answer Verification & ACC & In choice & - & test \\
\midrule
dart\_text\_generation & Data to Text & GPT & - & - & test \\
\midrule
drug\_extraction\_ade & Named Entity Recognition & F1 & In context & - & test \\
\midrule
scitail11\_sentence\_generation & Sentence Composition & ROUGE & - & - & test \\
\midrule
daily\_dialog\_formal\_classification & Dialogue Act Recognition & ACC & In choice & - & test \\
\midrule
smcalflow\_sentence\_generation & Dialogue Generation & ROUGE & - & - & test \\
\midrule
ethos\_text\_classification & Toxic Language Detection & ACC & In choice & - & test \\
\bottomrule
\end{tabular}
\caption{Details of 81-128 task in the SupNatInst dataset.}
\label{Tab:SupNatInst128}
\end{table*}

\begin{table*}
\centering
\scriptsize
\begin{tabular}{l|ccccc}
\toprule
\textbf{Name} & \textbf{Category} & \textbf{Metric} & \textbf{Scope} & \textbf{Format} & \textbf{Usage} \\
\midrule
sale\_relevance & Relevance & ACC & In choice & Json & train \\
\midrule
commodity\_alignment & Alignment & Match & In choice & List & train \\
\midrule
ingredient\_identification & Identification & F1 & In context & Json & train \\
\midrule
recommendation & Recommendation & ACC & In choice & - & train \\
\midrule
click\_prediction & Recommendation & ACC & In choice & Json + Explanation & train \\
\midrule
user\_interest\_mining & Mining & F1 & In choice & List & train \\
\midrule
recipe\_generation & Generation & BLEU & - & Step 1~2~3~ & train \\
\midrule
product\_description\_generation & Generation & BLEU & Contain center word & - & train \\
\midrule
summary\_generation & Generation & ROUGE & - & - & train \\
\midrule
food\_entity\_extraction & Named Entity Recognition & F1 & In context + In entity types & Json & train \\
\midrule
comment\_entity\_extraction & Named Entity Recognition & F1 & In context + In entity types & Json & train \\
\midrule
text2sql & Code & ACC & - & Legal sql & train \\
\midrule
merchant\_classification & Classification & ACC & In choice & - & train \\
\midrule
item\_classification & Classification & ACC & In choice & - & test \\
\midrule
logical\_reasoning & Reasoning & ACC & In choice & Uppercase letter & test \\
\midrule
conversation\_completion & Completion & BLEU & - & Length <= 50 & test \\
\midrule
**\_ner & Named Entity Recognition & F1 & In context + In entity types & Json & test \\
\midrule
property\_rel & Relevance & ACC & In choice & - & test \\
\midrule
post\_extraction & Named Entity Recognition & ACC & In context + In entity types & Json & test \\
\midrule
food\_rewrite & Rewriting & GPT & - & Length <= 7 & test \\
\bottomrule
\end{tabular}
\caption{Details of each task in the Domain dataset.}
\label{Tab:domain}
\end{table*}

\section{Templates}\label{template}
Table~\ref{Tab:template} shows our English templates for key-part extraction and instruction diversity.
We have four evolving strategies.
The strategies of concretizing, reasoning, and constraint make instructions more detailed, complex, and longer. 
The breadth strategy rewrites the general description within the instruction while keeping the key parts and length of the instruction nearly unchanged.

We obtain more combinations of key parts and instructions for each task with the following instruction-diversity process.
Initially, each task has an instruction pool, which contains a manually written instruction related to the task definition.
For each task, we first randomly select an instruction as the seed instruction from the instruction pool. 
We use the OpenAI-API\footnote{\url{https://api.openai.com/v1/chat/completions}} (gpt-4-0613\footnote{\url{https://platform.openai.com/docs/models/gpt-4-and-gpt-4-turbo}}) and key-part extraction template to extract key parts for the seed instruction. 
Then we randomly apply one strategy from the four evolving templates on the seed instruction to obtain the evolution instruction. 
Finally, we extract the key parts of the evolution instruction and add them to the instruction pool. 
We iteratively repeat such process until the size of the instruction pool reaches $31$.

\section{Hyperparameter Sensitivity}\label{algorithm}

\begin{figure}[h]
    \centering
    \includegraphics[width=\linewidth]{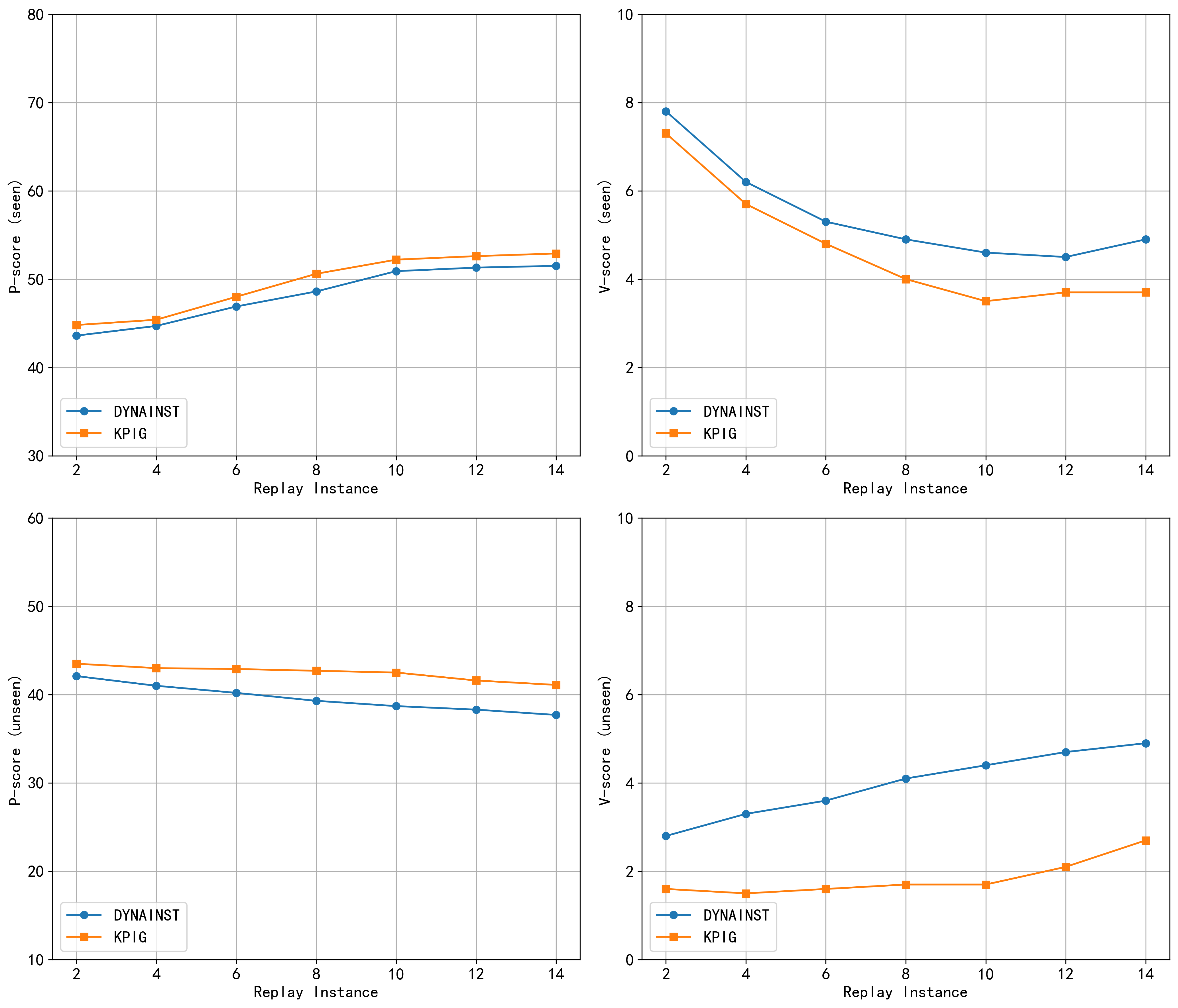}
    \caption{The impact of $N$ on model performance.}
    \label{fig:hyper}
\end{figure}

In our experiments, we finetune the exponential decay $\alpha$ in $\{0.05, 0.1, 0.3, 0.6, 1.0\}$, the weight $\lambda$ in $\{0.001$, $0.01$, $0.02$, $0.03$, $0.05$, $0.1\}$, 
the learning rate in $\{5e-6$, $1e-5$, $2e-5$, $3e-5$, $5e-5\}$ according to the average \texttt{P-score} on seen and held-out tasks.
In addition, since the number of replay instances $M$ and $N$ are key hyperparameters that affect model performance and runtime, we conduct further experiments on it.
As shown in Figure \ref{fig:hyper}, we report the influence of $N$, and find that increasing $N$ improves the \texttt{P-score} of seen tasks, but may have a negative effect on the performance of held-out tasks and instruction-following abilities. 
Therefore, we choose a trade-off reports based on the average \texttt{P-score} on seen and held-out tasks by fixing $M$ and then find $N$.

As shown in Table~\ref{tab:hyper}, we add more experiments on hyperparameter and conduct further analysis. When fixing 
$\alpha=0.3$, increasing $\lambda$
 from 0.001 to 0.1 shows a clear trade-off. While the P-score of seen tasks decreases from 54.7 to 47.6, the P-score of held-out tasks increases from 37.1 to 44.0, indicating stronger regularization benefits for generalization. Moreover, with fixed $\lambda=0.02$, increasing $\alpha$ from 0.3 to 1.0 causes the P-score of both seen and held-out tasks to decline, suggesting excessive 
 values impair overall performance. This may be because as the generation progresses, the posterior probability of the subsequent tokens increases, requiring a smaller $\alpha$ penalty. Therefore, to achieve a relative optimal balance, we choose $\lambda=0.02$ and $\alpha=0.3$
 based on the average P-score on seen and held-out tasks. However, we must admit that our exploration of $\lambda$ and $\alpha$ has limitations, because these two indicators may be related to the alignment tax or other issues of the CL process of the LLM. Dynamically adjusting these two hyperparams according to our IG may be more rigorous.

\begin{table}
\centering
\resizebox{\linewidth}{!}{
\begin{tabular}{ccc|ccc}
\hline
$\lambda$ ($\alpha=0.3$) & Seen P & Unseen P & $\alpha$ ($\lambda=0.02$) & Seen P & Unseen P \\
\hline
0.001 & 54.7 & 37.1 & 0.05 & 51.9 & 41.7 \\
0.01 & 52.6 & 40.8 & 0.1 & 51.6 & 42.1 \\
0.02 & 52.2 & 42.5 & 0.3 & 52.2 & 42.5 \\
0.05 & 49.6 & 43.3 & 0.6 & 51.3 & 41.4 \\
0.1 & 47.6 & 44.0 & 1.0 & 48.6 & 38.5 \\
\hline
\end{tabular}}
\caption{Performance on Sup-NatInst-ST dataset with varying $\lambda$ and $\alpha$.}
\label{tab:hyper}
\end{table}

\section{Implementation}\label{algorithm}

Our detailed algorithm implementation is shown in Algorithm~\ref{alg:algorithm}.
In practice, sampling new instruction for current seed instruction from the instruction-diversity cached pool (line 13) can serve as a preprocessing step.
And the IG results and outputs of $\mathcal{M}_{t-1}$ calculated during the replay stage (line 3-8) can be reused during the fine-tuning stage (line 12-19).
Therefore, in real fine-tuning, our method only has $\mathcal{M}_{t}$ in the graphics memory.

For the experimental settings of other baselines, we give priority to the hyperparameters reported in their paper. 
In particular, to be fair and reproduce the better performance of baselines, $M$ and $N$ of \textbf{DYNAINST} adopt the same settings as \textbf{KPIG}, and we set $M$ to be the number of historical tasks and $N$ to be the same as \textbf{KPIG} in \textbf{PCLL} and \textbf{DCL}.
For InsCL, we set the maximum replayed number to $200$ (consistent with their publication).

As for the training time, our KPIG takes $340$ minutes to complete the training on Sup-NatInst-ST dataset.
Compared with \textbf{SFT} ($200$ minutes), the extra time cost is mainly in the calculation stage of information gain, which takes about $1$ minutes for each time step.
In addition, under the setting of $M=10$ and $N=10$, \textit{replay}-based methods like \textbf{PCLL}, \textbf{DCL} and \textbf{DYNAINST} take about $300$ minutes, and our time difference ($40$ minutes) is that the logits of the masked part need further calculation.
\textbf{InsCL} takes about $440$ minutes due to the setting of a larger replay number.
However, to achieve the results reported in Table~\ref{Tab:main}, \textbf{PCLL} and \textbf{DCL} need to replay all historical tasks, which takes about $400$ minutes.
The above analysis shows that our method offers a relatively balanced trade-off between performance and training efficiency under the setting of CIT.

\begin{algorithm}[tb]
\caption{Algorithm of our proposed KPIG}
\label{alg:algorithm}
\textbf{Input}: A sequence of task sets $\mathcal{T}^T = \{\tau_{1}, \tau_{2}, \ldots,\tau_{n}\}_{t=1}^T$, initial foundation LLM $\mathcal{M}_{0}$, instruction-diversity cached pool $\mathcal{I}_{\tau}$ for each task $\tau$\\
\textbf{Output}: Target LLM $\mathcal{M}_{T}$\\
\begin{algorithmic}[1] 
\STATE $t \leftarrow t+1$
\WHILE{$t <= T$}
\STATE Replay task set $\mathcal{R}=\{\}$
\FOR {each $\tau \in \mathcal{T}^{t'<t}$}
\STATE Randomly sample $N$ instances and calculate IG for them $\triangleright {\rm Eq.}$ \ref{ig}
\STATE Calculate the average IG of $N$ instances as the IG of task $\tau$ via $\mathcal{M}_{t-1}$
\ENDFOR
\STATE Put the $M$ tasks with the lowest IG into $\mathcal{R}$, $|\mathcal{R}|=M \times N$ 
\STATE $\mathcal{T}^{t} \leftarrow \mathcal{T}^{t} \cup \mathcal{R}$
\STATE Deepcopy $\mathcal{M}_{t} \leftarrow \mathcal{M}_{t-1}$
\STATE Frozen $\mathcal{M}_{t-1}$
\FOR {each instance $d_\tau=(i^{seed}, c, y) \in \mathcal{T}^{t}$}
\STATE Sample an instruction $i$ for $d_\tau$ from $\mathcal{I}_{\tau}$
\STATE Mask the key parts in $i$ with \texttt{[MASK]} symbol to obtain $i^m$
\STATE $d_\tau \leftarrow (i, c, y)$, $d_\tau^m \leftarrow (i^m, c, y)$
\STATE Get output of $d_\tau$ via $\mathcal{M}_{t}$, and apply $\mathcal{L}_{CE}$ on it
\STATE Get outputs of $d_\tau^m$ via $\mathcal{M}_{t-1}$ and $\mathcal{M}_{t}$, and apply $\mathcal{L}_{JSD}$ on them
\STATE Optimize Loss $\triangleright {\rm Eq.}$ \ref{loss}
\ENDFOR
\STATE $t \leftarrow t+1$
\ENDWHILE
\end{algorithmic}
\end{algorithm}

\section{Dataset}\label{dataset}
Table~\ref{Statistics of datasets} shows the details of our datasets.
For Sup-NatInst, we have $40$ held-out tasks. and we select $100$ instances from each task based on Self-BLEU score and label distribution, which are used for evaluation.
For Domain, we have $7$ held-out tasks, and we select $500$ instances from each task for evaluation.
The difference between ST and SC settings lies in the time steps. 
The former trains a single task at each time step, while the latter trains all tasks of different category at different time step.

In addition, we list the details of each task in Table~\ref{Tab:SupNatInst40}, Table~\ref{Tab:SupNatInst80}, Table~\ref{Tab:SupNatInst128} and Table~\ref{Tab:domain}.
We mark the evaluation method, format constraints, and response range for each task based on manual annotation.
For example, \texttt{In choice} usually represents a classification task that must be selected from within the scope of instruction constraints.
\texttt{In context + In entity type} represents a combination constraint on named entity recognition tasks, which means that entities of the given type must be extracted from the given context.
Based on these manually annotated constraints, we can calculate \texttt{P-score} and \texttt{V-score} for all tasks.

\begin{table*}
\centering
\small
\begin{tabular}{p{\textwidth}}
\toprule
\multicolumn{1}{c}{\textbf{Ins$_1$}} \\
\midrule
\midrule
\texttt{\textbf{Instruction:} In this assignment, you are presented with a dialogue segment, a piece of communication between a User and an Agent. Your responsibility is to identify the speaker. The User generally instructs the Agent to arrange activities, obtain event details, and inspect the timetable. In contrast, the Agent's reply is a response to the User's queries or additional inquiries based on the User's directive. Respond with "User" or "Agent". Additionally, it's important to note that the User may also ask the Agent to cancel events.}\\
\texttt{\textbf{Key parts:} identify the speaker,
        User generally instructs the Agent to arrange activities, obtain event details, and inspect the timetable,
        Agent's reply is a response to the User's queries or additional inquiries based on the User's directive,
        Respond with "User" or "Agent",
        User may also ask the Agent to cancel events}\\
\midrule
\multicolumn{1}{c}{\textbf{Ins$_2$}} \\
\midrule
\midrule
\texttt{\textbf{Instruction:} For this activity, you are presented with an excerpt from a dialogue involving a User and an Agent. It is your task to identify who is speaking. The User typically instructs the Agent to organize events, obtain data on events, or survey the event plan. In contrast, the Agent's replies often address the User's queries or extend the conversation based on the User's directives. Please respond with either "User" or "Agent".}\\
\texttt{\textbf{Key parts:} identify who is speaking,
            User typically instructs the Agent to organize events, obtain data on events, or survey the event plan,
            Agent's replies often address the User's queries or extend the conversation based on the User's directives,
            Please respond with either "User" or "Agent"}\\
\midrule
\multicolumn{1}{c}{\textbf{Ins$_3$}} \\
\midrule
\midrule
\texttt{\textbf{Instruction:} In this task, you will be presented with a statement, a fragment of a dialogue between a User and an Agent. Your responsibility is to identify the speaker. The User typically instructs the Agent to organize events, gather details about events, and verify the schedule. Conversely, the Agent's reply is a response to the User's inquiries or additional queries based on the User's directive. Respond with either "User" or "Agent".}\\
\texttt{\textbf{Key parts:} identify the speaker,
            User typically instructs the Agent to organize events, gather details about events, and verify the schedule,
            Agent's reply is a response to the User's inquiries or additional queries based on the User's directive,
            "Respond with either "User" or "Agent"}\\
\midrule
\multicolumn{1}{c}{\textbf{Ins$_4$}} \\
\midrule
\midrule
\texttt{\textbf{Instruction:} In this task, you are presented with a dialogue fragment, a piece of conversation between a User and an Agent. Your responsibility is to identify the speaker. The User typically instructs the Agent to arrange events, fetch details about events, and verify the schedule. Conversely, the Agent's reply is a response to the User's inquiries or additional queries based on the User's directive. Respond with either "User" or "Agent".}\\
\texttt{\textbf{Key parts:} identify the speaker,
            User typically instructs the Agent to arrange events, fetch details about events, and verify the schedule,
            Agent's reply is a response to the User's inquiries or additional queries based on the User's directive,
            Respond with either "User" or "Agent"}\\
\midrule
\multicolumn{1}{c}{\textbf{Ins$_5$}} \\
\midrule
\midrule
\texttt{\textbf{Instruction:} In this task, you are presented with a dialogue fragment from a conversation between a User and an Agent. Your responsibility is to identify the speaker. The User typically instructs the Agent to organize events, fetch details about events, and verify the schedule. Conversely, the Agent's reply is a response to the User's inquiries or additional queries based on the User's directive. Respond with either "User" or "Agent".}\\
\texttt{\textbf{Key parts:} identify the speaker,
            User typically instructs the Agent to organize events, fetch details about events, and verify the schedule,
            Agent's reply is a response to the User's inquiries or additional queries based on the User's directive,
            Respond with either "User" or "Agent"}\\
\midrule
\multicolumn{1}{c}{\textbf{Ins$_6$}} \\
\midrule
\midrule
\texttt{\textbf{Instruction:} In this assignment, you are presented with a snippet of a dialogue between a User and an Agent. The User typically instructs the Agent to organise events, gather details about an event, and inspect the agenda, whilst the Agent's reply consists of answers to the User's inquiries or additional questions pertaining to the User's directive. Your task is to identify the speaker from the dialogue snippet, taking into consideration the typical role of the User and the Agent, and to provide the speaker's identity as "User" or "Agent". Additionally, ensure your judgement is supported by reasonable analysis of the given dialogue.}\\
\texttt{\textbf{Key parts:} identify the speaker from the dialogue snippet,
            taking into consideration the typical role of the User and the Agent,
            provide the speaker's identity as "User" or "Agent",
            ensure your judgement is supported by reasonable analysis of the given dialogue}\\
\bottomrule
\end{tabular}
\caption{Six held-out instructions and corresponding key parts of the smcalflow classification task.}
\label{Tab:appheld-out}
\end{table*}

\section{6 Held-out Instructions}

The six held-out instructions we used in our investigations on the information gain (\S\ref{Investigations on Information Gain}) are listed in Table~\ref{Tab:appheld-out}.
The smcalflow classification task is a seen task, which requires determining whether the sentence is spoken by a user or an agent.
Ins$_6$ has the smallest information gain and the worst model performance in Table~\ref{Tab:held_out}. 
This may be because it is not concise enough and has more redundant constraints compared with other instructions, which may indicate that our information gain may be helpful in measuring the clarity of the task definition.

\begin{figure}[htbp]
  \centering
  \subfigure[]{
    \includegraphics[width=0.45\linewidth]{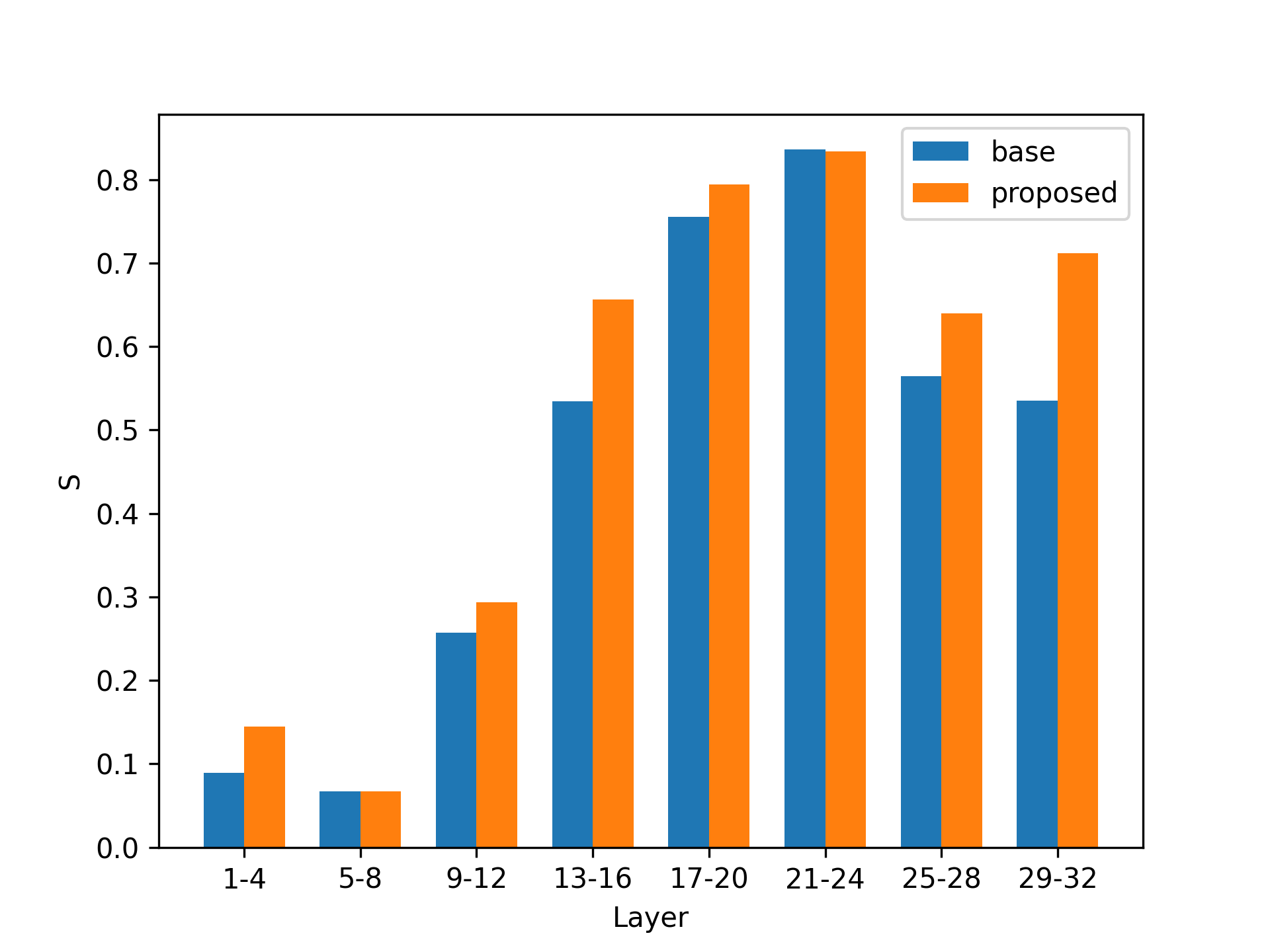}
  }
  \subfigure[]{
    \includegraphics[width=0.45\linewidth]{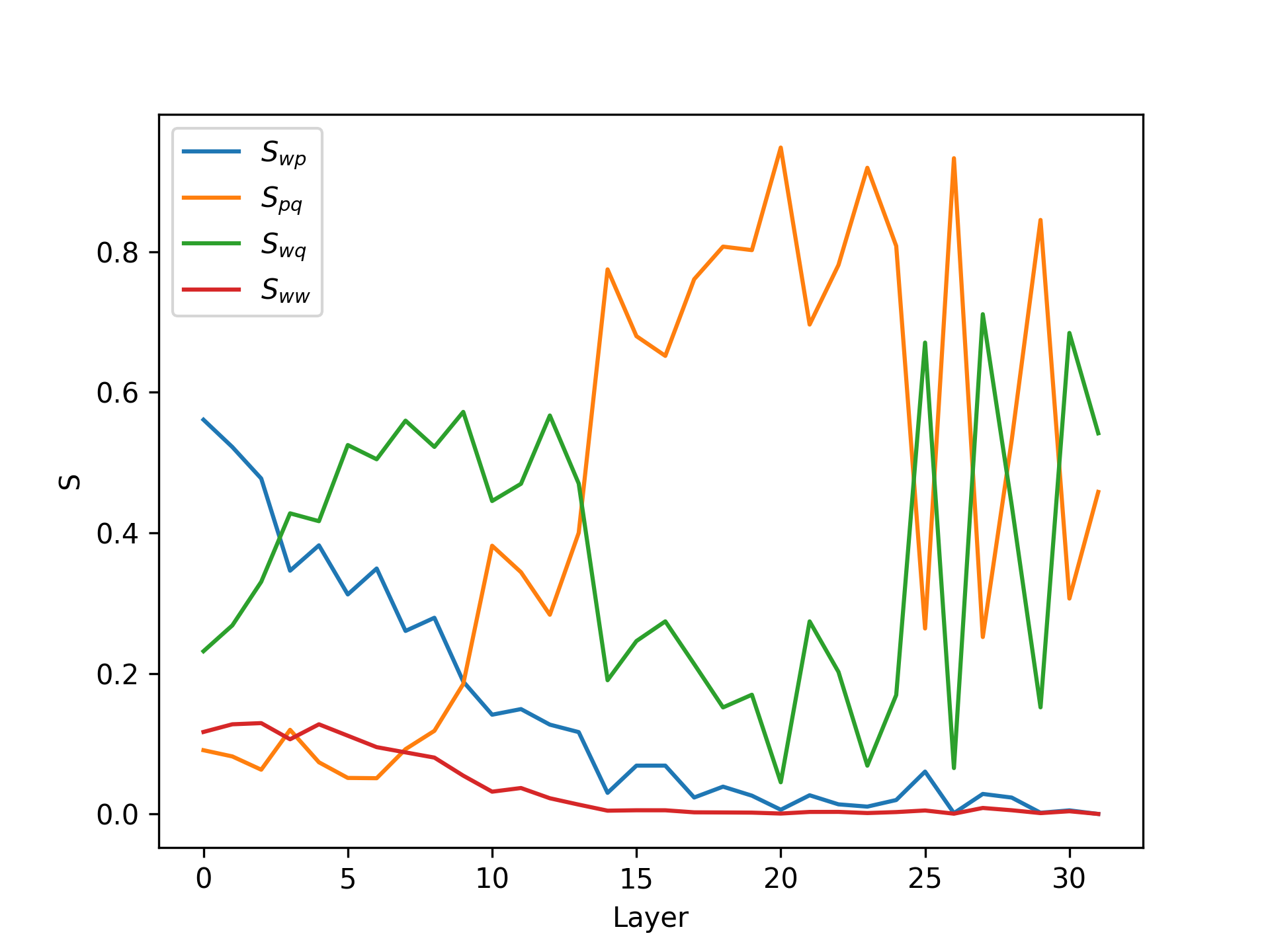}
  }
  \\
  \subfigure[]{
    \includegraphics[width=0.45\linewidth]{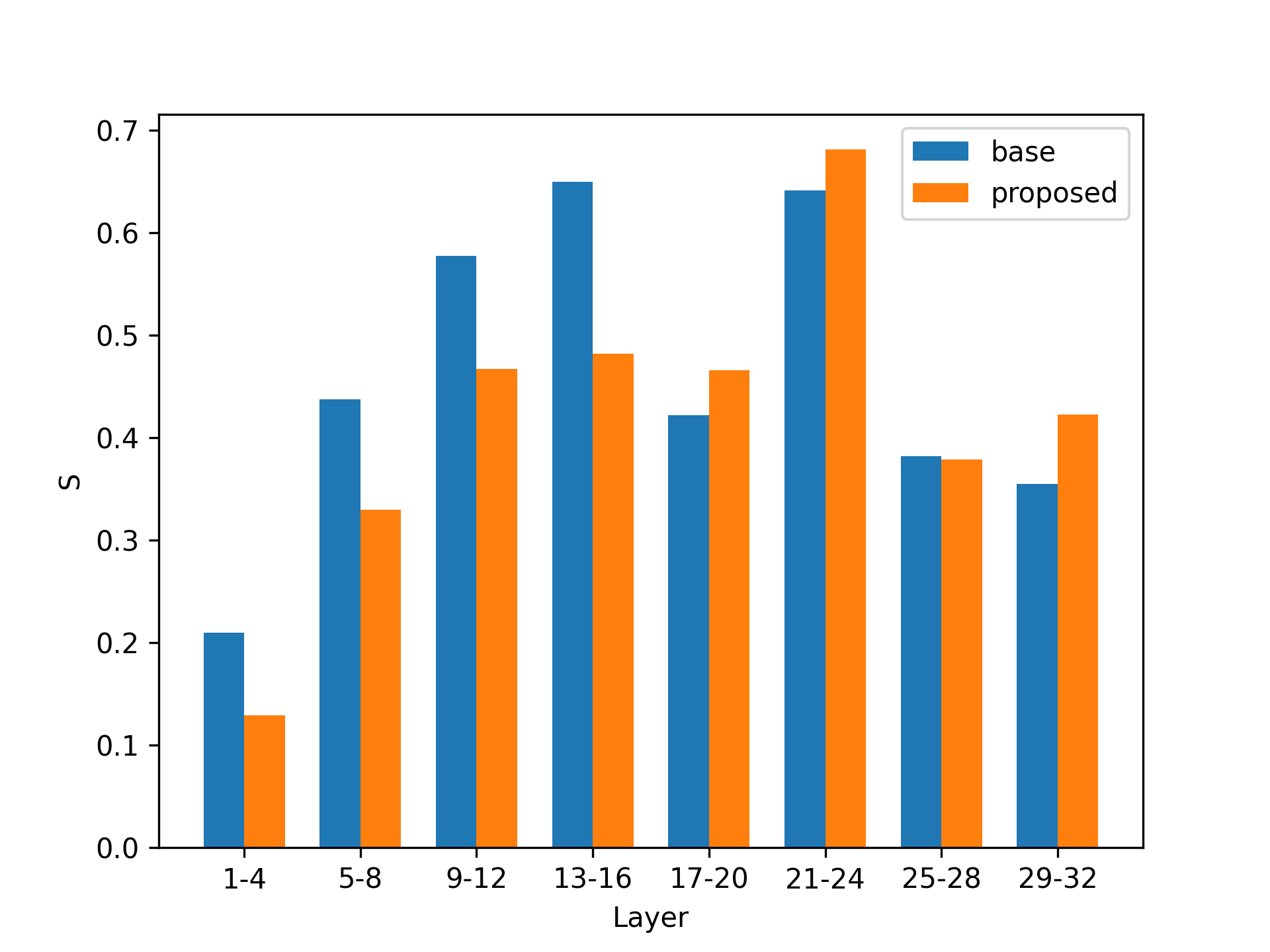}
  }
  \subfigure[]{
    \includegraphics[width=0.45\linewidth]{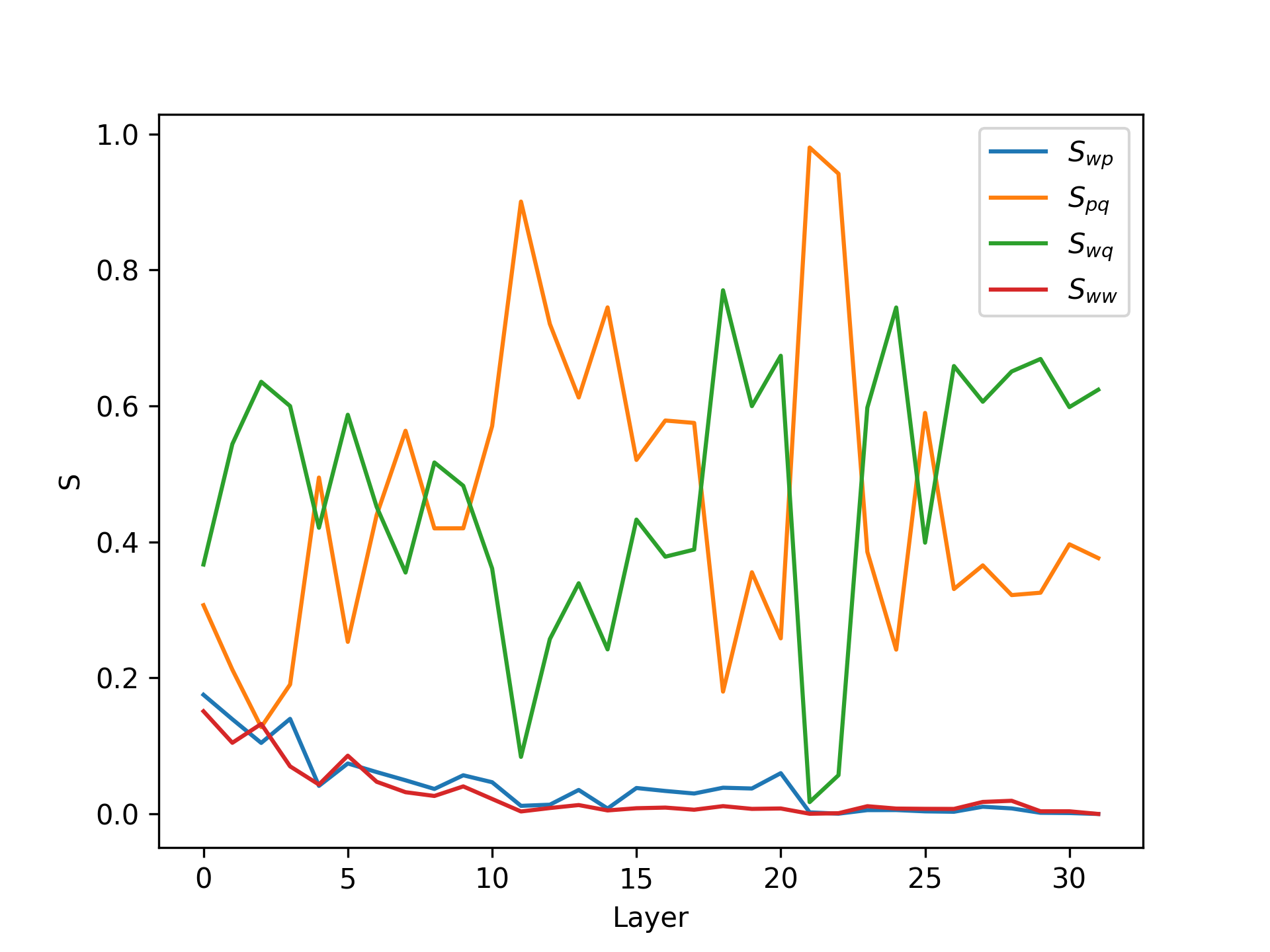}
  }
  \caption{Saliency scores of task 1645 and 1664.}
  \label{fig:all_images}
\end{figure}

\section{Investigations on Saliency Score}

We conduct further experiments on the saliency score \cite{DBLP:conf/emnlp/WangLDCZMZS23} of the ground-truth and key-part tokens. 
The saliency score is a crucial metric that indicates how much attention the model pays to relevant segments of user input. 
We randomly select two tasks from Sup-NatInst, which have significant constraints on the classification label space and the scope of generated content respectively.
As shown in Figure~\ref{fig:all_images}, ground truth tokens (e.g., 'S' in 'Similar' or 'D' in 'Dissimilar') receive a higher saliency score in the last 4 layers compared to baselines. 
For the INIT foundation LLM, attention flow towards the ground-truth tokens from the key parts is more significant than other parts across most tasks. 
The above results show that KPIG is more effective at identifying the key parts of user intent, leading to more accurate and reliable performance. 
However, compared to the classification-task 1645, such phenomenon is not so significant for generation-task 1664. This may be because the importance of context is not that different in such tasks, and it is difficult for GPT-4 to accurately identify the key parts (the instructions for such tasks have fewer constraints and definitions). 
And we believe these phenomena merit further exploration such as more carefully designed data construction and staged training.

\section{Details in Human Annotation}

In this section, we show the details of manual annotation on the constraints and the metric for each task. 
We recruited $4$ students aged 25 to 30 with computer background and proficient English communication skills. 
Since they are volunteers, they were not paid. 
We shuffled the data randomly and assigned data to them. 
The task is not included until at least $3$ people have consistent annotations. 
Our annotation instruction is like: "Given the instruction of the task definition, two positive demonstrations, two negative demonstrations, and the corresponding explanations, mark out the format (such as JSON, separator, upper and lower case, numbers, letters and so on), length restrictions, inclusion of specified words, selection from a specified range, and other constraints that are critical to generating desired responses. 
And choose the most applicable metric from F1, ACC, ROUGE, BLEU, Match, and GPT".
The above metrics are illustrated in \S\ref{3.4}, and we also give three demonstrations of applicable tasks for each metric.

\end{document}